\documentclass[journal]{IEEEtran}
\usepackage{blindtext}
\usepackage{graphicx}
\usepackage{wrapfig}

\usepackage{cite}

\ifCLASSINFOpdf
\else
\fi
\usepackage{amssymb}
\usepackage{amsfonts}
\usepackage[cmex10]{amsmath}
\DeclareMathOperator*{\argmin}{\arg\!\min}
\usepackage{url}

\newcommand*{\TitleFont}{%
      \usefont{\encodingdefault}{\rmdefault}{b}{n}%
      \fontsize{16}{22}%
      \selectfont}

\begin{document}
%
\title{\TitleFont Persistent self-supervised learning principle: \\from stereo to monocular vision for obstacle avoidance.}
%
%
%

\author{Kevin~van Hecke,
        Guido~de Croon,
        Laurens van der Maaten,        
        Daniel Hennes,
        and Dario~Izzo
\thanks{K.G. van Hecke and G.C.H.E. de Croon are with the MAV-lab, Control and Simulation, Faculty of Aerospace Engineering, Delft University of Technology, the Netherlands, L.J.P van der Maaten is with the Faculty of Computer Science, Delft University of Technology, the Netherlands, D. Hennes and D. Izzo are with the Advanced Concepts Team of the European Space Agency, ESTEC, Noordwijk, the Netherlands.
e-mail: k.g.vanhecke@tudelft.nl, g.c.h.e.decroon@tudelft.nl}
}

\setlength{\parindent}{0pt}

\maketitle

\begin{abstract}
Self-Supervised Learning (SSL) is a reliable learning mechanism in which a robot uses an original, trusted sensor cue for training to recognize an additional, complementary sensor cue. We study for the first time in SSL how a robot's learning behavior should be organized, so that the robot can keep performing its task in the case that the original cue becomes unavailable. We study this persistent form of SSL in the context of a flying robot that has to avoid obstacles based on distance estimates from the visual cue of stereo vision. Over time it will learn to also estimate distances based on monocular appearance cues. A strategy is introduced that has the robot switch from stereo vision based flight to monocular flight, with stereo vision purely used as `training wheels' to avoid imminent collisions. This strategy is shown to be an effective approach to the `feedback-induced data bias' problem as also experienced in learning from demonstration. Both simulations and real-world experiments with a stereo vision equipped AR drone 2.0 show the feasibility of this approach, with the robot successfully using monocular vision to avoid obstacles in a $5 \times 5$m room. The experiments show the potential of persistent SSL as a robust learning approach to enhance the capabilities of robots. Moreover, the abundant training data coming from the own sensors allows to gather large data sets necessary for deep learning approaches.

\end{abstract}


%
\IEEEpeerreviewmaketitle



\section{Introduction}
It is generally acknowledged that robots operating in the real world benefit greatly from learning mechanisms. Learning allows robots to adapt to environments or circumstances not specifically foreseen at design time. However, the outcome of learning and its influence on the learning robot's behavior can by definition not be predicted completely. This is a major reason for the delay in introducing successful learning methods such as Reinforcement Learning (RL) in the real world. For instance, with RL it is a major challenge to ensure an exploratory behavior that is safe for both the robot and its environment \cite{garcia2015comprehensive}. 

Learning from demonstration (LfD) can in this respect be regarded as more reliable. However, in the case of a mobile robot, LfD faces a `feedback-induced data bias' problem \cite{Ross2011,Ross2013a}. If the robot executes its trained policy on real sensory inputs, its actions will be slightly different from the expert's. As a result, the trajectory of the robot will be different to when the human expert was in control, leading to a test data distribution that is different from the training distribution. This difference worsens the performance of the learned policy, further increasing the discrepancy between the data distributions. The solution proposed in \cite{Ross2011,Ross2013a} is to have the human expert provide novel training data for the sensory inputs experienced by the robot when being in control itself. This leads to an iterative process that requires quite a time investment of the human expert.

There is a relatively new learning mechanism for robots that combines reliability with the advantage of not needing any human supervision. \emph{Self-Supervised Learning (SSL)} does not learn a control policy as LfD and RL, but rather focuses on improving the sensory inputs used in control. Specifically, in SSL the robot uses the outputs of an original, trusted sensor cue to learn to recognize an additional, complementary sensor cue. The reliability comes from the fact that the robot has access to the trusted cue during the entire learning process, ensuring a baseline performance of the system. 

Until now, the purpose of SSL has mostly been the exploitation of the complementarity between the sensor cues. To illustrate, perhaps the most well-known example is the use of SSL on Stanley, the car that won the grand DARPA challenge \cite{thrun2007stanley}. Stanley used a laser scanner to detect the road ahead. The range of the laser scanner was rather limited, which placed a considerable restriction on the robot's driving speed. SSL was used in order to extend the road detection beyond the range of the laser scanner. In particular, the available laser scanner based road classifications were used to train a color model of drivable terrain in camera images. This color model was then applied to image regions not covered by the laser scanner. These image regions higher up in the image allowed to detect the road further away. The use of SSL permitted Stanley to speed up considerably and was an important factor in winning the competition. 

A characterizing feature of many SSL studies \cite{lieb2005adaptive,thrun2007stanley,lookingbill2007reverse,Hadsell2009b,muller2013real}, is that the learning and SSL estimation takes place restricted to one moment of time, i.e. on one image in case of a video stream. This implies that the supervisory signal always has to be active as new images arrive from the stream. Only a few, very recent, studies have the learned function last longer over time \cite{baleia2015exploiting,ho2015optical}. The idea in these studies is to give the robot the capability to sometimes act based solely on the complementary cue. For instance, in \cite{baleia2015exploiting} the sense of touch is used to teach a vision process how to recognize traversable paths through vegetation  with as goal to gradually reduce time-intensive haptic interaction. Hence, the learning of recognizing the complementary cue will have to persist in time. However, the consequences of this persistent form of SSL on the robot's behavior when using the complementary cue have not been addressed in the above-mentioned studies. 

In this article, we perform an in-depth study of the behavioral aspects of persistent SSL in the context of a scenario in which the robot should keep performing its task even when the supervisory cue becomes completely unavailable. Importantly, when the robot relies only on the complementary cue, it will encounter the feedback-induced data bias problem known from LfD. We suggest a novel decision strategy to handle this problem in persistent SSL. 

Specifically, we study a flying robot with a stereo vision system that has to avoid obstacles. The robot uses SSL to learn a mapping from monocular appearance cues to the stereo-based distance estimates. We have chosen this context, because it is relevant for any stereo-based robot that needs to be robust to a permanent failure of one of its cameras. In computer vision monocular distance estimation is also studied. There,  the main challenges are the gathering of sufficient data (e.g., for deep neural networks) and the generalization of the learned distance estimation to an unforeseen operation environment. Both of these challenges are addressed to some extent by SSL, as learning data is abundant and the robot learns in the environment in which it operates. We regard SSL as an important supplement to machine learning for robots. Therefore we end the study with a discussion on the position of (persistent) SSL in the broader context of robot and machine learning, comparing it among others with reinforcement learning, learning from demonstration and supervised learning.


The remainder of the article is set up as follows. First, in Section \ref{section:related_work}, we discuss related work. Then, in Section \ref{section:methodology}, we more formally introduce persistent SSL and explain our implementation for the stereo-to-mono learning. We analyze the similarity of the specific SSL case studied in this article with Learning from Demonstration approaches. Subsequently, in Section \ref{section:offline} we perform offline vision experiments in order to assess the performance of various parameter settings. Thereafter, in Section \ref{section:simulation} we compare various learning strategies. The best learning strategy is implemented for experiments with a Parrot AR drone 2, the results and analysis of which are described in Section \ref{section:robot_experiments}. The broader implications of the findings on persistent SSL are discussed in Section \ref{section:discussion}, and conclusions are drawn in Section \ref{section:conclusion}.


\section{Related work} \label{section:related_work}
We study persistent SSL in the context of a stereo-vision equipped robot that has to learn to navigate with a single camera. In this section, we discuss the state-of-the-art in the most relevant areas to the study: monocular navigation and self-supervised learning.

\subsection{Monocular navigation}
The large majority of monocular navigation approaches focuses on motion cues. The optical flow of world points allows for the detection of obstacles \cite{Mori2013} or even the extraction of structure from motion, as in monocular Simultaneous Localization And Mapping (SLAM) \cite{Engel2014}. The main issue of using monocular motion cues for navigation is that optical flow only conveys information on the ratio between distance and the camera's velocity. Additional information is necessary for retrieving distance. This information is typically provided by additional sensors \cite{engel2014scale}, but can also be retrieved by performing specific optical flow maneuvers \cite{van2014monocular,de2016monocular}.

In contrast, it is well known that the appearance of objects in a single, still image does contain distance information. Successfully estimating distances in a single image allows robots to evaluate distances without having to move. In addition, many appearance extraction and evaluation methods are computationally efficient. Both these factors can aid the robot in the making of quick navigation decisions. Below, we give an overview of work in the area of monocular appearance-based distance estimation and navigation.

\subsubsection{Appearance-based navigation without distance estimation}
There are some appearance-based navigation methods that do not involve an explicit distance estimate. For instance, in \cite{DeCroon2012b}, an appearance variation cue is used to detect the proximity to obstacles, which is shown to be successful at complementing optical flow based time-to-contact estimates. A threshold is set that makes the flying robot turn if the variation drops too much, which will lead to turns at different distances. 

An alternative approach is to directly learn the mapping from visual inputs to control actions. In order to fly a quad rotor through a forest, Ross et al \cite{Ross2013a} use a variant of Learning from Demonstration (LfD) \cite{Argall2009} to acquire training data on avoiding trees in a forest. First a human pilot controls the drone, creating a training data set of sensory inputs and desired actions. Subsequently, a control policy is trained to mimick the pilot's commands as good as possible. This control policy is then implemented on the drone. 

A major problem of this approach is the \emph{feedback-induced data bias}: A robot has a feedback loop of actions and sensory inputs, so its control policy determines the distribution of world states that it encounters (with corresponding sensory inputs and optimal actions). Small deviations between the trained controller and the human may bring the robot in unknown states in which it has received no training. Its control policy may generalize badly to such situations. The solution proposed in \cite{Ross2013a} is a transitional model called DAgger \cite{Ross2011}, in which actions from the expert are mixed with actions from the trained controller. In the real-world experiments in \cite{Ross2013a}, several iterations have been performed in which the robot flies with the trained controller, and the captured videos are labeled offline by a human. This approach requires skilled pilots and significant human effort. 


\subsubsection{Offline monocular distance learning}
Humans are able to see distances in a single still image, and there is a growing body of work in computer vision utilizing machine learning to do the same. Interest in single image depth estimation was sparked by work from Hoiem et al \cite{Hoiem2005} and Saxena et al \cite{Saxena2007a}, \cite{Saxena2009a}. Hoiem's Automatic Photo Pop-up tries to group parts of the image into segments that can be popped up at different depths. Saxena's Make-3D uses a Markov Random Field (MRF) approach to classify a depth per image patch on different scales. These studies focus on creating a dense depth map with a machine learning computer vision approach. Both methods use supervised learning on a large training dataset. Some work was done on adopting variants of Saxena's MRF work for driving rovers and even for MAVs. Lenz et al \cite{Lenz2012} propose a solution based on a MRF to detect obstacles on board an MAV, but it does not infer how far the objects are. Instead, it is trained offline to recognize three different obstacle class types. Any different object could hence lead to navigation problems. 

Recently, again focusing on creating a dense depth map from a single image, Eigen et al \cite{Eigen2014} propose a multi-scale deep neural network approach trained on the KITTI dataset, making it more resilient for practical robot data. Training deep neural networks requires a large data set, which is often obtained by deforming training data or by artificially generating training data. Michels et al \cite{Michels2005} use artificial data to learn recognizing obstacles on a rover, but in order to generalize well it requires the use of a very realistic simulator. In addition, the same work reports significant improvement if the artificial data is augmented with labeled real-world data. 

Other groups acquire training data for supervised learning by having another separate robot or system acquire data. This data is then processed and learned offline, after which the learned algorithm is deployed on the target robot. Dey et al. \cite{Dey} use an RC car with a stereo vision setup to acquire data from an environment, apply machine learning on this data offline, and navigate a similar but unseen environment with an MAV based on the trained algorithm. Creating and operating a secondary system designed to acquire training data, however, is no free lunch. Moreover, it introduces inconvenient biases in the training data, because an RC car will not behave in a similar way both in terms of dynamics, camera viewpoint and the path chosen through the environment.

None of the above methods have the robot gather the data and learn while in operation.

\subsection{Self-supervised learning}
The idea of self-supervised learning has been around since the late 1990s \cite{Yamauchi1999}, but the successful application of it to terrain classification on the autonomously driving car Stanley \cite{Thrun2007a} demonstrated its first major practical use. A similar approach was taken by Hadsel et al. \cite{Hadsell2009b}, but now using a stereo vision system instead of a LIDAR system, and complex convolutional filters instead of simple and fixed color based features. These approaches largely forgo the need for manually labeled data as they are designed to work in unseen environments. 

In the studies on self-supervised learning for terrain classification, the ground truth is assumed to be always available. In fact, the learned function only lasts for a single image, implying that the trained terrain appearance models are directly applicable to very similar data.  

 
 Two very recent studies do have the learned function last longer over time, both with the idea of having the robot take some decisions based on the complementary sensor cue alone. Beleia et al \cite{baleia2015exploiting} study a rover with a haptic antenna sensor. In their application of terrain mapping they try to map monocular cues to obstacles based on earlier events of encountering similar situations which resulted in either a hard obstacle, a traversable obstacle, or a clear path. The monocular information is used in a path planning task, requiring a cost function for either exploring unknown potential obstacles or driving through a terrain on the current available information. Since checking whether a potential obstacle is traversable is costly (the rover needs to travel there in order for the antenna to provide ground truth on that), the robot learns to classify the terrain ahead with vision. On each sample an analysis is performed to determine whether the vision-based classifier is sufficiently confident: it either decides the terrain is traversable, not traversable, or unsure. In the unsure case, the sample is sensed using the antenna. Gradually this will become necessary less often, thus learning to navigate using its Kinect sensor alone. In Ho et al. \cite{ho2015optical}, a flying robot first uses optical flow to select a landing site that is flat and free of obstacles. In order for this to work, the robot has to move sufficiently with respect to the objects on the landing site. While flying, the robot uses SSL to learn a regression function that maps an (appearance-based) texton-distribution to the values coming from the optical flow process. The learned function extends the capabilities of the robot, as after learning it is also able to select landing sites without moving (from hover). The article shows that if the robot is uncertain on its appearance-based estimates, it can switch back to the original optical flow based cue. 
 
In this article, we focus on the behavioral aspects of persistent SSL. We study how to  best set up the learning process, so that the robot will be able to keep performing its task when the original sensor cue becomes completely unavailable.


\section{Methodology overview} \label{section:methodology}
This section starts by formally and generally posing the persistent SSL learning mechanism (Subsection \ref{section:PSSL_principle}). Next, in Subsection \ref{section:stereo_to_mono}, we show how this method is used for our specific proof of concept case, monocular depth estimation in flying robots, providing subsequent implementation details in the following subsections \ref{section:stereo_processing}-\ref{section:performance}. Lastly, in Subsection \ref{section:similarity_imitation_learning}, we identify a behavioral issue when applying persistent SSL and formally define the learning problem for our proof of concept. For a comparison between persistent SSL and other machine learning methods, we refer the reader to the discussion.

\subsection{Persistent SSL principle} \label{section:PSSL_principle}
\begin{figure}[htp]
\centering
\includegraphics[width=8cm]{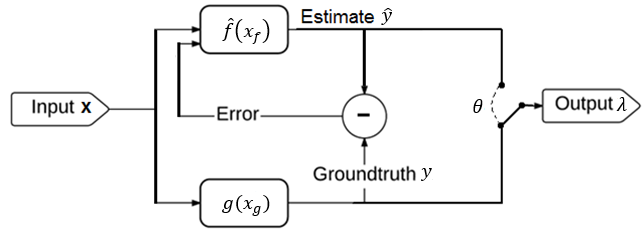}
\caption{The persistent self-supervised learning principle.}
\label{fig:Schematic}
\end{figure}

The persistent SSL principle is schematically depicted in Figure \ref{fig:Schematic}. In persistent SSL, an original, pre-wired sensory cue provides supervised outputs to a learning process that takes a different, complementary sensory cue as input. The goal is to be able to replace the pre-wired cue if necessary. When considering the system as a whole, learning with persistent SSL can be considered as unsupervised; it requires no manual labeling or pre-training before deployment in the field. Internally it uses a supervised learning method that in fact needs ground truth labels to learn. This ground truth is, however, assumed to be provided online and autonomously without human or outside interference.\smallskip

In the schematic, the input variable $\mathbf{x}$ represents the sensory inputs available on board. The variables $x_g$ and $x_f$ are possibly overlapping subsets of these sensory inputs. In particular, function $g(x_g)$ extracts a trusted, ground truth sensory cue from the sensory inputs $x_g$. In classical systems, $g(x_g)$ provides the required functionality on its own:
\begin{equation}
g:x_g \rightarrow y, \; \; \; \; x_g \subseteq \mathbf{x}
\label{eq:g}
\end{equation}

The function $f(x_f)$ is learned with a supervised learning algorithm in order to approximate $g(x_g)$ based on $x_f$: 
\begin{equation}
f:x_f \rightarrow \hat{y}, \; \; \; \; f \in F, \; x_f \subseteq \mathbf{x}
\label{eq:f}
\end{equation}
\begin{equation} \label{eq:pSSL1}
\hat{f} = \argmin_{f \in F} \mathbb{E} \left[ l(f(x_f), g(x_g))  \right],
\end{equation}
\noindent where $l(f(x_f), g(x_g))$ is a suitable loss function that is to be minimized. The system can either choose or be forced to switch $\theta$, so that $\lambda$ is either set to $g(x_g)$ or $\hat{f}(x_f)$ for use in control. Future work may also include fusing the two, but in this article we focus on using the complementary cue in a stand-alone fashion. It must be noted that while both $x_g\subseteq x$ and $x_f\subseteq x$, in general it may be that  $x_f$ does not contain all necessary information to predict $y$. In addition, even if $x_g = x_f$, it is possible that $F$ does not contain a function $f$ that perfectly models $g$. The information in $x_f$ and the function space $F$ may not allow for a perfect estimate of $g(x_g)$. On the other hand, there may be an $f(x_f)$ that handles certain situations better than $g(x_g)$ (think of landing site selection from hover, as in \cite{ho2015optical}). In any case fundamental differences between $g(x_g)$ and $\hat{f}(x_f)$ are to be expected, which may significantly influence the behavior when switching $\theta$. Handling these differences is of central interest in this article. 

\subsection{Stereo-to-mono proof of concept} \label{section:stereo_to_mono}
\begin{figure*}
\centering
\includegraphics[trim={0cm 8cm 5cm 0cm},clip,width=16cm]{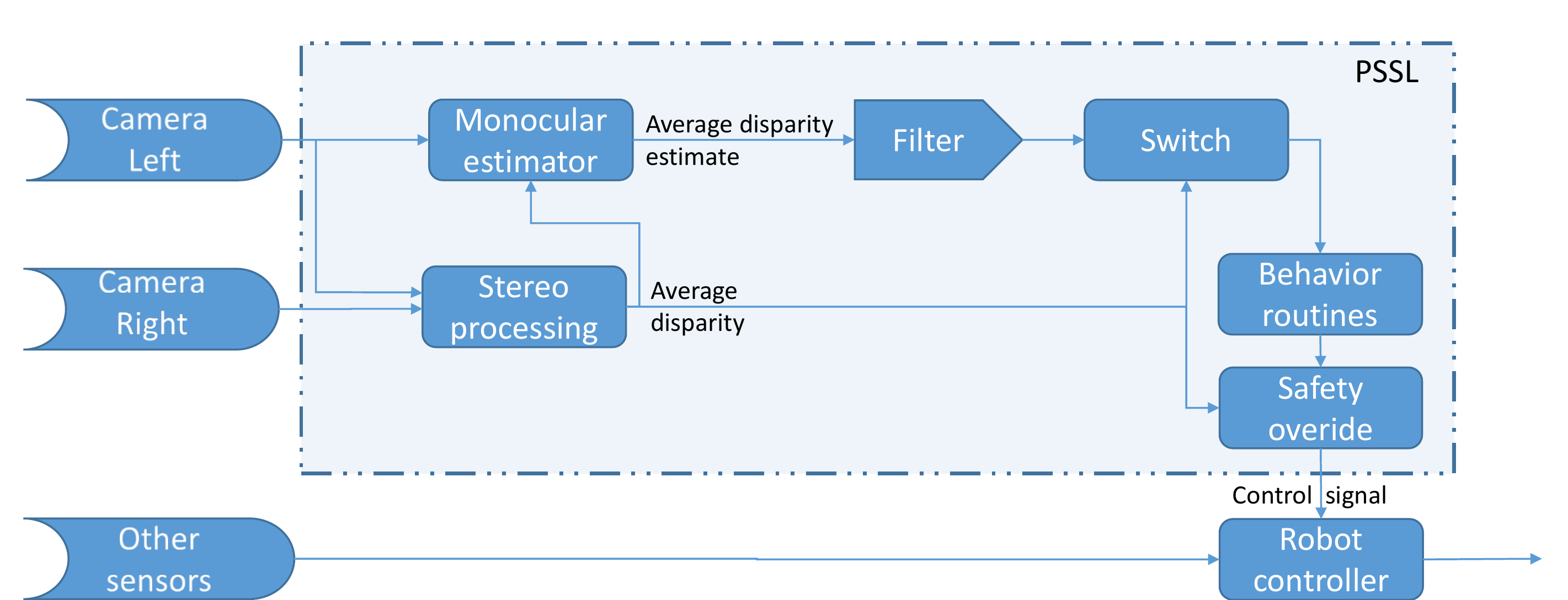}
\caption{System overview. Please see the text for details.}
\label{fig:Overview}
\end{figure*}

Figure \ref{fig:Overview} presents a block diagram of the proposed proof of concept system in order to visualize how the persistent SSL method is employed in our application: estimating monocular depth in a flying robot. 
Input is provided by a stereo vision camera, with either the left or right camera image routed to the monocular estimator. We use a Visual Bag of Words (VBoW) method for this estimator (see Subsection \ref{section:mono}). The ground truth for persistent SSL in this context is provided by the output of a stereo vision algorithm. In this case, the average value of the disparity map is used, both for training the monocular estimator and as an input to the switch $\theta$. Based on the switch, the system either delivers the monocular or the stereo vision average disparity to the behavior controller. 


\subsection{Stereo vision processing} \label{section:stereo_processing}
The stereo camera delivers a synchronized gray-scale stereo-pair image per time sample. A stereo vision algorithm first computes a disparity map, but often this is a far from perfect process. Especially in the context of an MAV's size, weight and computational constraints, errors caused by imperfect stereo calibration, resolution limits, etc. can cause large pixel errors in the results. Moreover, learning to estimate a dense disparity map, even when this is based on a high quality and consistent data set, is already very challenging. Since we use the stereo result as ground truth for learning, we minimize the error by averaging the disparity map to a single scalar. A single scalar is much easier to learn than a full depth map and has been demonstrated to provide elementary obstacle avoidance capability \cite{Wagter2014a}, \cite{Michels2005}, \cite{DeCroon2012c}.\smallskip

The disparity $\lambda$ relates to the depth $d$ of the input image:
\begin{equation} 
d \propto \frac{1}{\lambda}
\label{eq:disparity}
\end{equation}
Using averaged disparity instead of averaged depth fits the obstacle avoidance application better, because small but close by objects are emphasized due the non-linear relation of Eq. \ref{eq:disparity}. However, linear learning methods may have difficulty mapping this relation. In our final design we thus choose to learn the disparity with a non-parametric approach, which is resilient to nonlinearities.

\subsection{Monocular disparity estimation} \label{section:mono}
The monocular disparity estimator forms a function from the image's pixel values to the average disparity in the image. Since the main goal of the article is to study SSL on board a drone in real-time, we have put efficiency of both the learning and execution of this function are at a prime. Hence, we converged to a computationally extremely efficient Visual Bag of Words (VBoW) approach for the robotic experiments. We have also explored a deep neural network approach, but the hardware and time available for learning did not allow for having the deep neural learning on board the drone at this stage. 

\begin{figure}[htp]
\centering
\includegraphics[width=8cm]{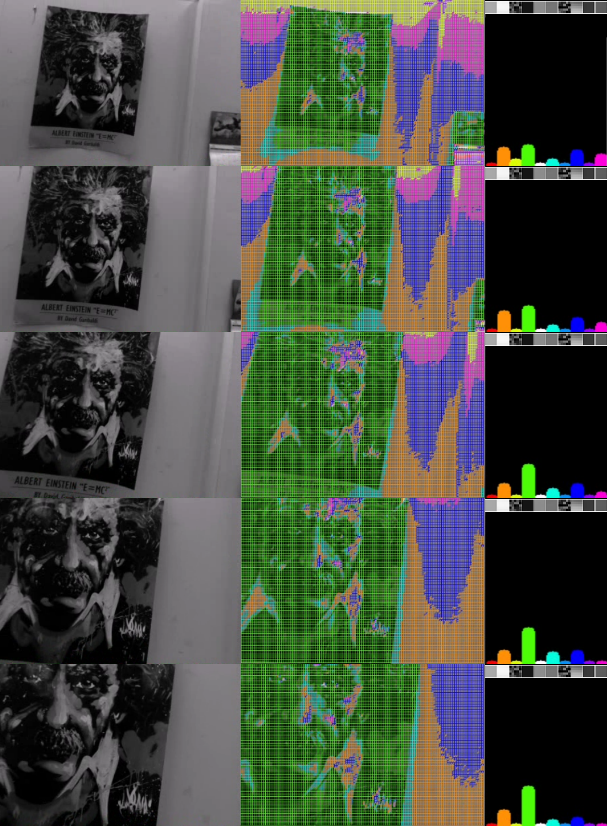}
\caption{Approaching a poster on the wall. Left: monocular input. Middle: overlaid textons annotated with the color used in the histogram. Right: texton distribution histogram with the corresponding texton shown beneath it.}
\label{fig:Closing}
\end{figure}

The VBoW method uses small image patches of $w \times h$ pixels, as successfully introduced in \cite{Varma2003} for a complex texture classification problem. First, a dictionary is created by clustering the image patches with Kohonen clustering (as in \cite{DeCroon2012c}). The $n$ cluster centroids are called `textons'. After formation of the dictionary, when an image is received, $m$ patches are extracted from the $W \times H$ pixel image. Each patch is compared to the dictionary in order to form a texton occurrence histogram for the image. The histogram is normalized to sum to 1. Then, each normalized histogram is supplemented with its Shannon entropy, resulting in a feature vector of size $n+1$. The idea behind adding the entropy is that the variation of textures in an image decreases when the camera gets closer to obstacles \cite{DeCroon2012c}. To illustrate the change in texton histograms when approaching an obstacle, a time series of texton histograms can be seen in Figure \ref{fig:Closing}. Please note how the entropy of the distribution indeed decreases over time, and that especially the fourth bin is much higher when close to the poster on the wall. A machine learning algorithm will have to learn to notice such relationships itself for the robot's environment, by learning a mapping from the feature vector to a disparity. We have investigated different function representations and learning methods to this end (see Section \ref{section:offline}).

\subsection{Behavior} \label{section:behavior}
\begin{figure}[htp]
\centering
\includegraphics[trim={0.0cm 9.5cm 26cm 0.8cm},clip,width=5cm]{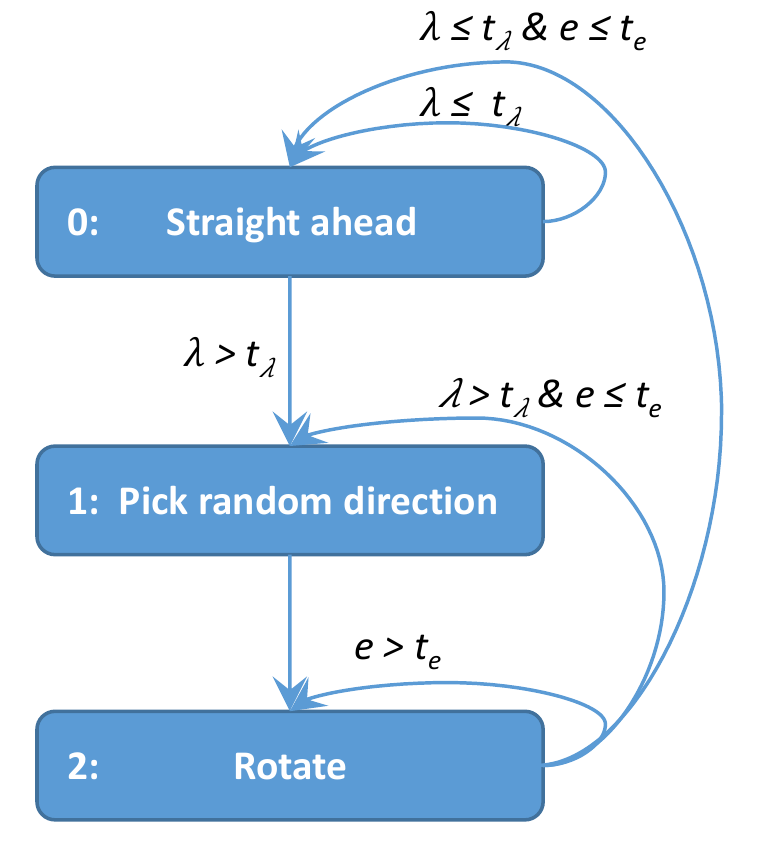}
\caption{The behavior heuristic FSM. $\lambda$ is average disparity, $e$ is the attitude error (meaning the difference between the newly picked direction and the current attitude), $t_n$ the respective thresholds.}
\label{fig:BehaviorHeuristic}
\end{figure}

The proposed system uses a straightforward behavior heuristic to explore, navigate and persistently learn a room. The heuristic is depicted as a Finite State Machine (FSM) in Figure \ref{fig:BehaviorHeuristic}. In state 0 the robot flies in the direction of the camera's principal axis. When an obstacle is detected ($\lambda > t$), the robot stops and goes to state 1 in which it randomly chooses a new direction for the principal axis. It immediately passes to state 2 in which the robot rotates towards the new direction, reducing the error $e$ between the principal axis' current and desired direction. If in the new direction obstacles are far enough away ($\lambda \leq t$), the robot starts flying forward again (state 0). Else, the robot continues to turn in the same direction as before (clockwise or counter clockwise) until $\lambda \leq t$. When this is the case, it starts flying straight again (state 0). The FSM detects obstacles by means of a threshold $t$ applied to the average disparity $\lambda$. Choosing this rather straightforward behavior heuristic enables autonomous exploration based on only one scalar obtained from a distance sensor. 


\subsection{Performance} \label{section:performance}
The average disparity $\lambda$, coming either from stereo vision or from the monocular distance estimation function $f(x_f)$, is thresholded for determining whether to turn. This leads to a binary classification problem, where all samples for which $\lambda > t$ are considered as `positive' ($c=1$) and all samples for which $\lambda \leq t$ are considered as `negative' ($c=0$). Hence, the quality of $\hat{f}(x_f)$ can be characterized by a ROC curve. The ground truth for the ROC curve is determined by the stereo vision. This means that a True Positive Ratio (TPR) of 1 and False Positive Ratio (FPR) of 0 lead to the same obstacle detection performance as with the stereo vision system. Generally, of course, this performance will not be reached, and the robot has to determine what threshold to set for  a sufficiently high TPR and low FPR. 


This leads to the question how to determine what a `sufficient' TPR / FPR is. We evaluate this matter in the context of the robot's obstacle avoidance task. In particular, we first look at the probability of a collision with a given TPR and then at the probability of a spurious turn with a given FPR.

In order to model the probability of a collision, consider a constant velocity approach with input samples (images) $ \langle x_1, x_2, \ldots, x_n \rangle$ of $n$ samples long, ending at an obstacle. A minimum of $u$ samples before actual impact, an obstacle must be detected by at least one TP or a FP in order to prevent a collision. Since the range of samples $\langle x_{(n-u+1)}, x_{(n-u+2)}, \ldots, x_n \rangle$ does not matter for the outcome, we redefine the approach range to be $\langle x_1, x_2, \ldots, x_{(n-u)}\rangle$. Consider that for each sample $x_i$ holds:
\begin{equation}
1 = p(TP \vert x_i) + p(FP \vert x_i )+p(TN \vert x_i )+p(FN \vert x_i ),
\label{eq:TFPN}
\end{equation}
\noindent since  $p(FN \vert x_i ) = p(TP \vert x_i) = 0$ if $x_i$ is a negative and $p(TN \vert x_i ) = p(FP \vert x_i) = 0$ if $x_i$ is a positive. Let us first assume independent, identically distributed (i.i.d.) data. Then, the probability of a collision $p_c$ can be written as:
\begin{equation}
\begin{split}
p_{c} & = \prod_{i=1}^{n-u} \left( p(FN \vert x_i ) + p(TN \vert x_i ) \right) \\
      & = \prod_{i=1}^{q} p(TN \vert x_i ) \prod_{i=q+1}^{n-u} p(FN \vert x_i ), \\
\label{eq:pcollision}
\end{split}
\end{equation}
\noindent where $q$ is a time step separating two phases in the approach. In the first phase all $x_i$ are negative, so that any false positive will lead to a turn, preventing the collision. Only if all negative samples are correctly classified as negatives (true negatives), will the robot enter the second phase in which all $x_i$ are positive. Then only a complete sequence of false negatives will lead to a collision, since any true positive will lead to a timely turn.

We can use Eq. \ref{eq:pcollision} to choose an acceptable $TPR = 1 - FNR$. Assuming a constant velocity and frame rate, it gives us the probability of a collision. For instance, let us assume that the robot flies forward at $0.50$m/s with a frame rate of $30$Hz, it has a minimal required detection distance of $1.0$m and positives are defined to be closer than $1.5$m. This leads to $s=30$ samples that all have to be classified as negatives. In the case of i.i.d. data, if the $TPR = 0.95$, the probability of a collision is $p_c = (1-TPR)^s = 0.05^{30} \approx 9.31 \: 10^{-40}$, an extremely small value. With this analysis, even a $TPR=0.30$ leads to an acceptable $p_c \approx 2.25 \: 10^{-5}$. 

The analysis of the effect of false positives is straightforward, as it can be expressed in the number of spurious turns per second or, equivalently if assuming a constant velocity, per meter travelled. With the same scenario as above, an $FPR = 0.05$ will on average lead to $3$ spurious turns per travelled meter, which is unacceptably high. An $FPR=0.0017$ will approximately lead to $1$ spurious turn per 10 meters. 

The above analysis seems to indicate that quite many false negatives are acceptable, while there can only be very few false positives. However, there are two complicating factors. The first factor is that Eq. \ref{eq:pcollision} only holds when $X$ can be assumed identically and independently distributed (i.i.d.), which is unlikely due to the nature of the consecutive samples of an approach towards an obstacle. Some reasoning about the nature of the dependence is, however, possible. Assuming a strong correlation between consecutive samples results in a higher probability of $x_i$ being classified the same as $x_{(i+1)}$. In other words, if a sample in the range $\langle x_1..x_{m}\rangle$ is an FN, the chance that more samples are FNs increases. Hence, the expected dependencies negatively impact performance of the system making Eq. \ref{eq:pcollision} a best case scenario. 

\begin{figure*}
\centering
\includegraphics[trim={0cm 4cm 0cm 0cm},clip,width=16cm]{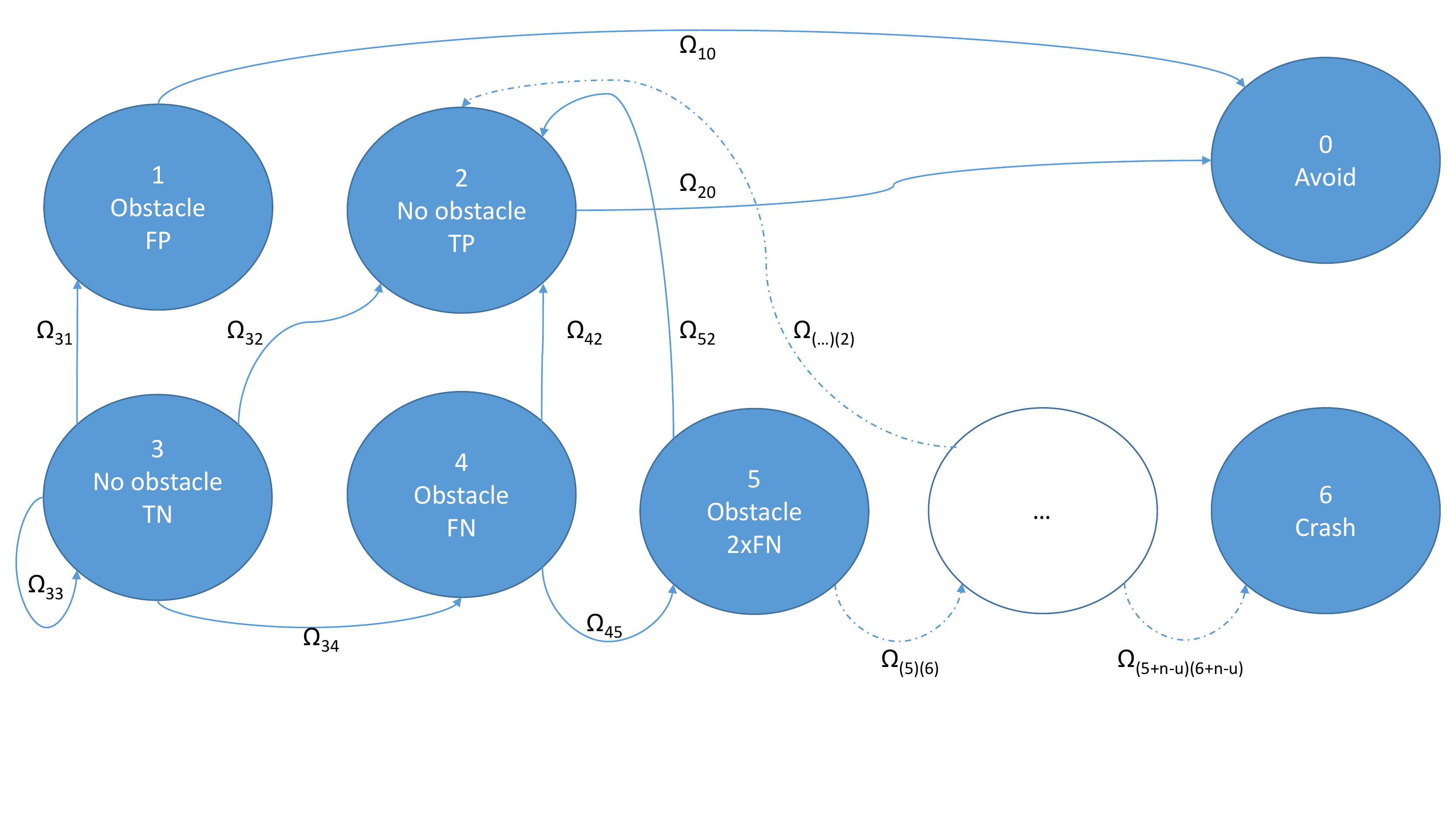}
\caption{Markov model of the probability of a collision.}
\label{fig:MarkovModel}
\end{figure*}

The system can be more realistically modelled as a Markov process as depicted in Figure \ref{fig:MarkovModel}. From this can be seen that the system can be split in a reducible Markov process with an absorbing  \textit{avoid} state, and a chain of states that leads to the absorbing  \textit{collision} state. The values of the transition matrix $\Omega$ can be determined from the data gathered during operation. This would allow the robot to better predict the consequences of a chosen $TPR$ and $FPR$. 

As an illustration of the effects of sample dependence, let us suppose a model in which each classification has a probability of being identical to the previous classification, $p(I(c_{i-1}, c_{i}))$. If not identical, the sample is classified independently. This dependency model allows us to calculate the transition $\Omega_{4,5}$ in Figure \ref{fig:MarkovModel}. Given a previous negative classification, the transition probability to another negative classification is: $\Omega_{4,5}= p(I(c_{i-1}, c_{i})) + (1-p(I(c_{i-1}, c_{i}))) (1-TPR)$. If $p(I(c_{i-1}, c_{i})) = 0.8$ and $TPR = 0.95$ as above, $\Omega_{4,5}=0.81$. The probability of a collision in such a model is $p_c = \Omega_{4,5}^{(s-1)} = 1.8 \: 10^{-3}$, no longer an inconceivably small number. 

This leads us to the second complicating factor, which is specific to our SSL setup. Since the robot operates on the basis of the ground-truth, it should theoretically hardly ever encounter positive samples. Namely, the robot should turn when it detects a positive sample. This implies that the uncertainty on the estimated TPR is rather high, while the FPR can be estimated better. A potential solution to this problem is to purposefully have the mono-estimation robot turn earlier than the stereo vision based one.


\subsection{Similarity with Learning from Demonstration} \label{section:similarity_imitation_learning}
The core of SSL is a supervised algorithm that learns the function $\hat{f}(x_f)$ on the basis of supervised outputs $g(x_g)$. Normally, supervised learning assumes that the training data is drawn from the same data probability distribution $\mathcal{D}$ as the test data. However, in persistent SSL this assumption generally does not hold. The problem is that by using control based on $\hat{f}$, the robot follows a control policy $\pi_{\hat{f}} \neq \pi_{g}$ and hence will induce a different state distribution, $\mathcal{D}_{\pi_{\hat{f}}} \neq \mathcal{D}_{\pi_g}$. On these different states, no supervised outputs have been observed yet, which typically implies an increasing difference between $\hat{f}$ and $g$. 

A similar problem of inducing a different state distribution is well-known in the area of learning from demonstration \cite{Ross2011,Ross2013a}. Actually, we will show that under some mild assumptions, the persistent SSL problem studied in this paper is equivalent to an learning from demonstration problem. Hence, we can draw on solutions in learning from demonstration such as DAgger \cite{Ross2011}. 

The goal of learning from demonstration is to find a policy $\hat{\pi}$ that minizes a loss function $l$ under its induced distribution of states, from \cite{Ross2011}:
\begin{equation} \label{eq:IL}
\hat{\pi} = \argmin_{\pi \in \Pi} \mathbb{E}_{s \sim \mathcal{D}_{\pi}} \left[ l(s, \pi) \right],
\end{equation}
\noindent where an optimal, teacher policy $\pi^*$ is available to provide training data for specific states $s$.

At first sight, SSL is quite different, as it focuses only on the state information that serves as input to the policy. Instead of optimizing a policy, the supervised learning in  persistent SSL can be defined as finding the function $f'$ that best matches the trusted function $g'$ under the distribution of states induced by the use of the thresholded version $f$ for control:
\begin{equation} \label{eq:pSSL}
\argmin_{f \in F} \mathbb{E}_{x \sim \mathcal{D}_{\pi_f}} \left[ l(f(x), g(x))  \right] ,
\end{equation}
\noindent meaning that we perform regression of $f$ on states that are induced by the control policy $\pi_f$, which uses the thresholded version of $f$. 

To see the similarity to Eq. \ref{eq:IL}, first realize that the stereo-based policy is in this case the teacher policy: $\pi_g = \pi^*$. For this analysis we simplify the strategy to flying straight when far enough away from an obstacle, and turning otherwise:
\begin{equation}
\begin{split}
\pi_g(s): \quad & p(\mathrm{straight} | s=0 ) = 1 \\
          & p(\mathrm{turn} | s=1) = 1, \\
\end{split}
\end{equation}
\noindent where $s$ is the state, with $s=1$ when $g(x) > t_g$ and $s=0$ otherwise. Please note that $\pi_g$ is a deterministic policy, which is assumed to be optimal.

When we learn a function $\hat{f}$, it generally will not give exactly the same outputs as $g$. Using $\hat{s} := \hat{f} > t_{\hat{f}}$ will result in the following stochastic policy:
\begin{equation}
\begin{split}
\pi_{\hat{f}}(s): \quad & p(\mathrm{straight} | s=0) = TNR \\
                  & p(\mathrm{turn} | s=0) = FPR \\
                  & p(\mathrm{turn} | s=1) = TPR \\
                  & p(\mathrm{straight} | s=1) = FNR, \\
\end{split}
\end{equation}
\noindent a stochastic policy which by definition is optimal, $\pi_{\hat{f}} = \pi_g$, if $FPR = FNR = 0$. In addition, then $\mathcal{D}_{\pi_{\hat{f}}} = \mathcal{D}_{\pi_g}$. Thus, if we make the assumption that minimizing $l(f(x), g(x))$ also minimizes $FPR$ and $FNR$, capturing any preference for one or the other in the cost function for the behavior $l(s, \pi)$, then minimizing $f$ in Eq. \ref{eq:pSSL} is equivalent to minimizing the loss in Eq. \ref{eq:IL}. 

The interest of the above-mentioned similarity lies in the use of proven techniques from the field of learning from demonstration for training the persistent SSL system. For instance, in DAgger, during the first learning iteration only the expert policy is learned. After this iteration, a mix of the learned and expert policy is used: $\pi_i = \pi^*$, with $p = \beta_i$ and $\pi_i = \hat{\pi_i}$ with $p = (1-\beta_i)$, where $\beta_i$ is reduced over the iterations\footnote{In \cite{Ross2011} this mixture was written as $\pi_i = \beta_i \pi^* + (1-\beta_i) \hat{\pi}$, hinting at a mixed control. However, it is described as a policy \emph{that queries the expert to choose controls a fraction of the time while collecting the next dataset}. For this reason, and because it makes more sense given our binary control strategy, we mention this policy as a probabilistic mixture in this article.}. The policy at iteration $i$ depends on all previous learned data, which is aggregated over time. In \cite{Ross2011} it is proven that this strategy gives a favorable limited no-regret loss, which significantly outperforms the traditional learning from demonstration strategy of just learning from the data distribution of the expert policy. In Section \ref{section:simulation} we will show that the main lessons from \cite{Ross2011} also apply to persistent SSL.


\section{Offline Vision Experiments} \label{section:offline}
In this section, we perform offline vision experiments. The goal of these experiments is to determine how good the proposed VBoW method is at estimating monocular depth, and to determine the best parameter settings. 

To measure the performance, we use two main metrics: the mean square error (MSE) and the area under the curve (AUC) of a ROC curve. MSE is an easy metric that can be directly used as a loss function, but in practice many situations exist in which a low MSE can be achieved, but still inadequate performance is reached for the basis of reliable MAV behavioral control. The AUC captures the trade-off between TPR and FPR and hence is a good indication of how good the performance is in terms of obstacle detection. 

\begin{figure}[htp]
\centering
\includegraphics[width=8cm]{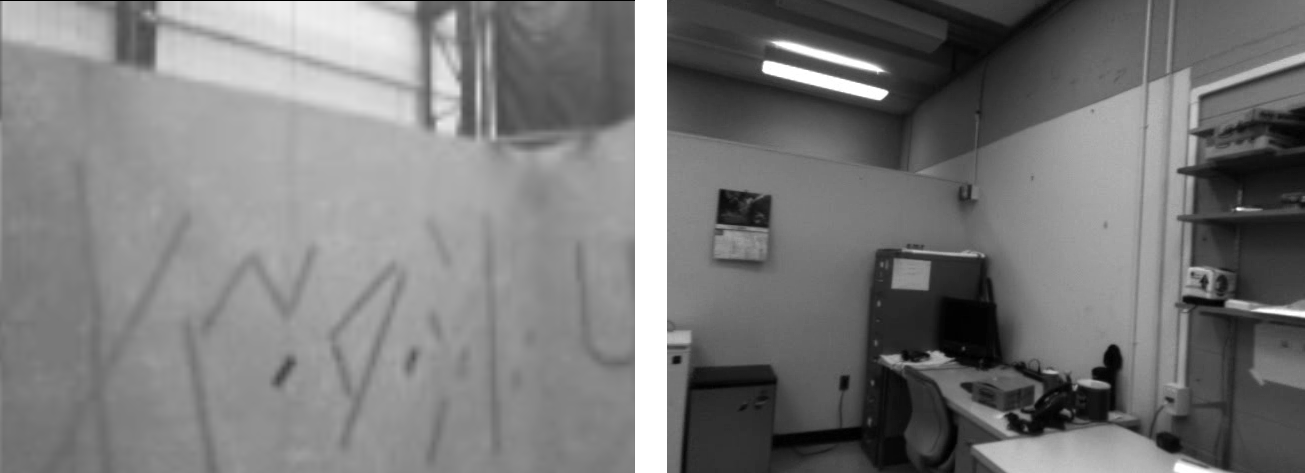}
\caption{Example from Dataset \#1 (left, 128x96 pixels) and dataset \#2 (right, 640x480).}
\label{fig:Rooms_dataset}
\end{figure}

We use two data sets in the experiments. The first dataset is a video made on a drone during an autonomous flight using the onboard $128 \times 96$ pixels stereo camera. The second dataset is a video made by manually walking with a higher quality $640 \times 480$ pixel stereo camera through an office cubicle in a similar fashion as the robot should move in the later online experiments. The datasets \#1 and \#2 used in this section are made available for download publicly\footnote{Dataset \#1 can be downloaded and viewed from: \url{http://1drv.ms/1NOhIOh}}\footnote{Dataset \#2 can be downloaded from: \url{http://1drv.ms/1NOhLtA}}. An example image from each dataset is shown in Figure \ref{fig:Rooms_dataset}.

Our implementation of the VBoW method has six main parameters, ranging from the number of intensity and gradient textons to the number of samples used to smooth the estimated disparity over time. An exhaustive search of parameters being out of reach, we have performed various investigations of parameter changes along a single dimension. Table \ref{table:1} presents a list of the final tuned parameter values. Please note that these parameter values have not only been optimized for performance. Whenever performance differences were marginal, we have chosen the parameter values that saved on computational effort. This choice was guided by our goal to perform the learning on board of a computationally limited drone. Below we will show a few of the results when varying a single parameter, deviating from the settings in Table \ref{table:1} in the corresponding dimension. 

\begin{figure}[htp]
\centering
\includegraphics[width=8cm]{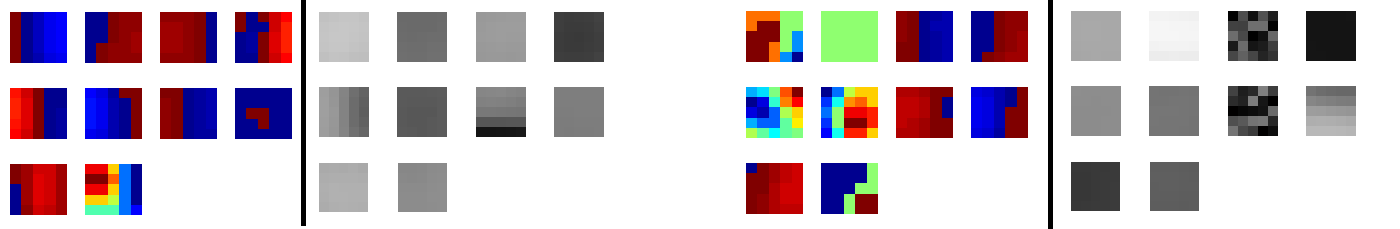}
\caption{Used texton dictionaries. Left for camera \#1, right for \#2.}
\label{fig:Dictionaries}
\end{figure}

In this work two types of textons are combined to form a single texton histogram: normal intensity textons as obtained with Kohonen clustering as in \cite{DeCroon2012c} and gradient textons obtained similarly but based upon the gradient of the images. Gradient textures have been shown in \cite{Wu2004} to be an important depth cue. An example dictionary of each is depicted in figure \ref{fig:Dictionaries}. Gradient textons are shown with a color range (from blue = low, to red = high). The intensity textons in figure \ref{fig:Dictionaries} are based on grayscale intensity pixel values. 


\begin{figure*}[htp]
\centering
\includegraphics[trim={1cm 6cm 3 5cm},clip,width=12cm]{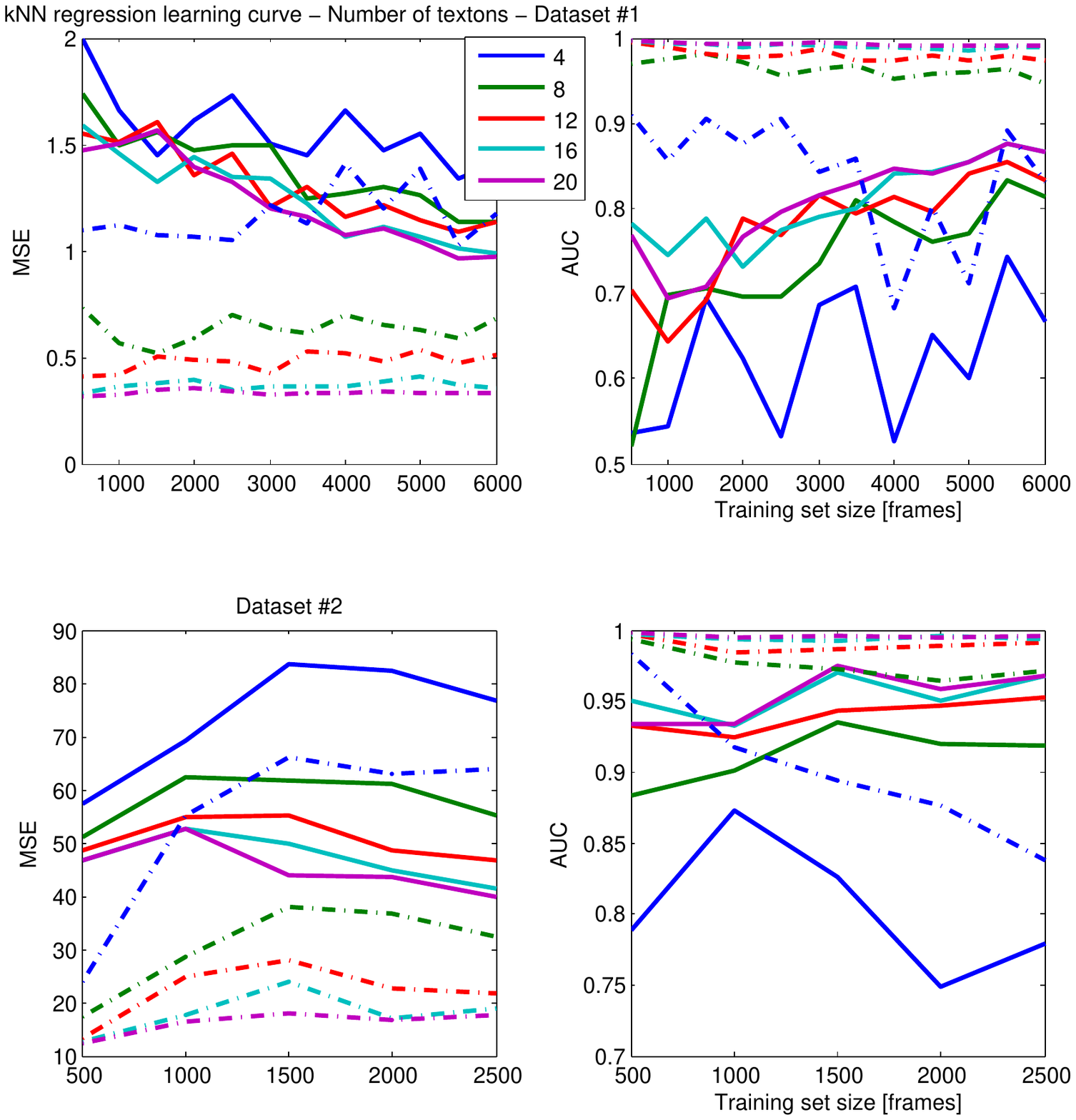}
\caption{MSE and AUC  number of textons. Dashed/ solid lines refer to results on train/test set.}
\label{fig:learnc_nTextons}
\end{figure*}

Figure \ref{fig:learnc_nTextons} shows the results for different numbers of textons, $\in \{ 4, 8, 12, 16, 20\}$, always consisting of half pixel intensity and half gradient textons. From the results we can see that the performance saturates around 20 textons. Hence we selected this combination of 10 intensity and 10 gradient textons for the experiments.





\begin{figure*}[htp]
\centering
\includegraphics[trim={1cm 6cm 3 5cm},clip,width=12cm]{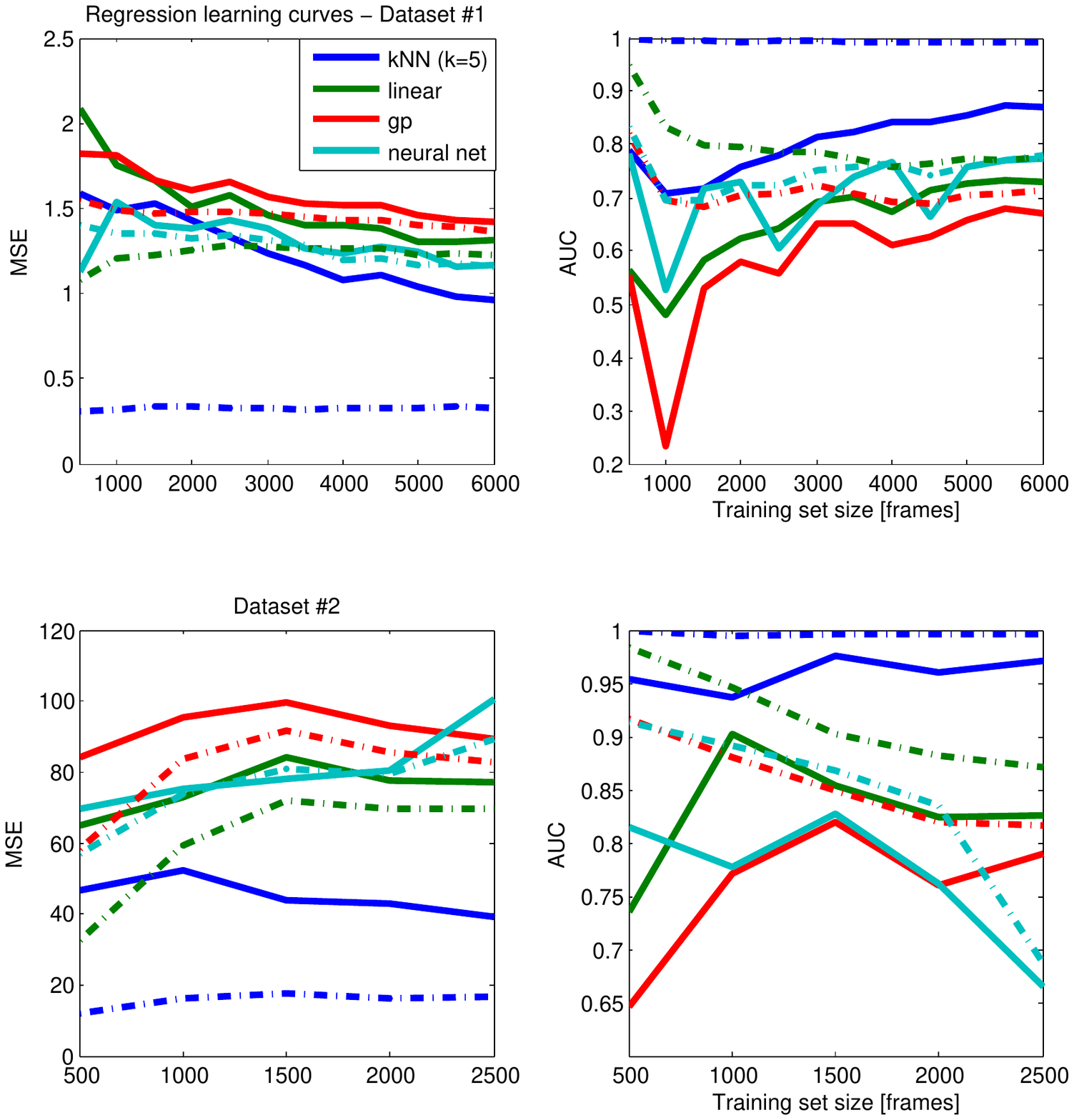}
\caption{VBoW regression algorithms learning curves. Dashed/ solid lines refer to results on train/test set.}
\label{fig:VBoWRegressors}
\end{figure*}

The VBoW method involves choosing a regression algorithm. In order to determine the best learning algorithm we have tested four regression algorithms, limiting the choice mainly based on feasibility for implementing the regression algorithm on board a constrained embedded system. We have tested two non-parametric (kNN and Gaussian Process regression) and two parametric (linear and shallow neural network regression) algorithms. Figure \ref{fig:VBoWRegressors} presents the learning curves for a comparison of these regressors. Clearly, in most cases the kNN regression comes out best. A na\"ive implementation of kNN suffers from having a larger training set in terms of CPU usage during test time, but after implementation on the drone this did not become a bottleneck.\smallskip

 The final offline results on the two datasets are quite satisfactory. They can be viewed online\footnote{A video of VBoW visualizations on dataset \#1 is available here: \url{http://1drv.ms/1KdRtC1}}\footnote{A video of VBoW visualizations on dataset \#2 is available here: \url{http://1drv.ms/1KdRxlg}}. After a training set of roughly $4000$ samples, the kNN approximates the stereo vision based disparities in the test set rather well. Given a desired TPR of $0.96$, the learner has an FPR of $0.47$. This should be sufficient for usage of the estimated disparities in control. 
 
\begin{table}[htb]
\caption{Parameter settings}
\centering
\begin{tabular}{||c|c||} 
 \hline
 Parameter & Value \\ [0.5ex] 
 \hline\hline
 Number of intensity textons & 10 \\ 
 \hline
 Number of gradient textons & 10 \\ 
 \hline
 Patch size & 5x5 \\
 \hline
 Subsampling samples & 500 \\
 \hline
 kNN & k=5 \\
 \hline
 Smooth size & 4 \\ [1ex] 
 \hline
\end{tabular}
\label{table:1}
\end{table}


\section{Simulation Experiments} \label{section:simulation}
In Section \ref{section:methodology}, we argued that a persistent form of SSL is similar to learning from demonstration. The relevance of this similarity lies in the behavioral schemes used for learning. In this section, we compare three learning schemes in simulation. 

\subsection{Setup}
\begin{figure}[htp]
\centering
\includegraphics[width=6cm]{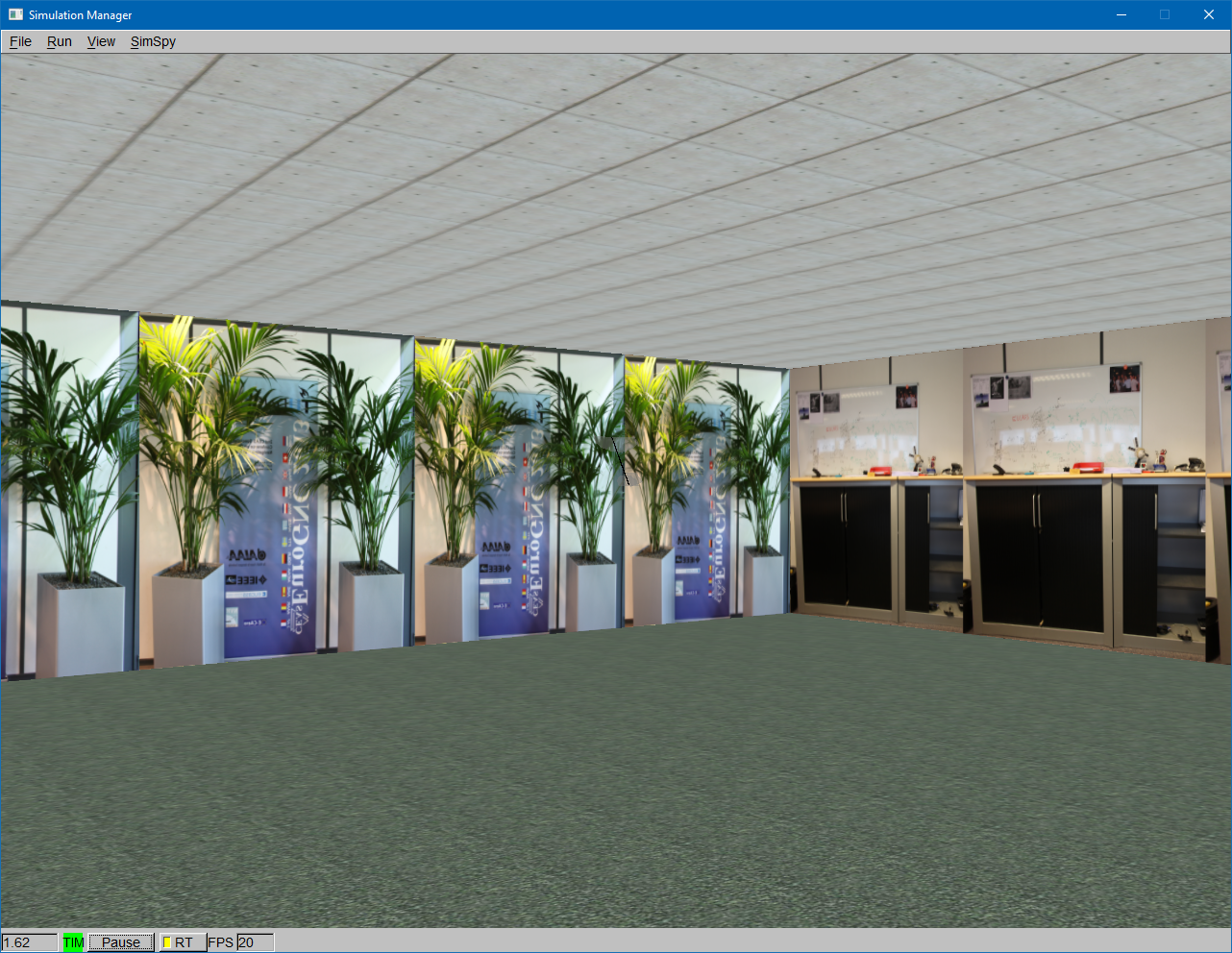}
\caption{SmartUAV simulation environment}
\label{fig:smartuav_3d}
\end{figure}

We simulate a `flying' drone with stereo vision camera in SmartUAV \cite{de2007holiday50av}, an in-house developed simulator that allows for 3D rendering and simulation of the sensors and algorithms used on board the real drone. Figure \ref{fig:smartuav_3d} shows the simulated `office room'. The room has a size of $10 \times 10$ meter, and the drone an average forward speed of $0.5$ m/s. All the vision and learning algorithms are exactly the same as the ones that run on board of the drone in the real experiments.

Three learning schemes are designed as follows. They all start with an initial learning period in which the drone is controlled purely by means of stereo vision. In the first learning scheme, the drone will continue to fly based on stereo vision for the remainder of the learning time. After learning, the drone immediately switches to monocular vision. For this reason, the first scheme is referred to as `cold turkey'. In the second learning scheme, the drone will perform a stochastic policy, selecting the stereo vision based actions with a probability $\beta_i$ and monocular based actions with a probability $(1-\beta_i)$, as was proposed in the original DAgger article \cite{Ross2011}. In the experiments, $\beta_i=0.25$. Finally, in the third learning scheme, the drone will perform monocular based actions, with stereo vision only used to override these actions when the drone gets too close to an obstacle. Therefore, we refer to this scheme as `training wheels'. 

After learning, the drone will use its monocular disparity estimates for control. The stereo vision remains active only for overriding the control if the drone gets too close to a wall. During this testing period, we register the number of turns and the number of overrides. The number of overrides is a measure of the number of potential collisions. The number of turns is compared to the number of turns performed when using stereo vision to evaluate the number of spurious turns. The initial learning period is $1$ minute, the remaining learning period is $4$ minutes, and the test time is $5$ minutes. These times have been selected to allow a full experiment on a single battery of the real drone (see Section \ref{section:robot_experiments}).

\subsection{Results}
Table \ref{tab:results_simulation} contains the results of 30 experiments with the three learning schemes and a purely stereo-vision-controlled drone. The first observation is that `cold turkey' gives the worst results. This result was to be expected on the basis of the similarity between persistent SSL and learning from demonstration: the learned monocular distance estimates do not generalize well to the test distribution when the monocular vision is in control. The originally proposed DAgger scheme performs better, while the third learning scheme termed `training wheels' seems most effective. The third scheme has the lowest number of overrides of all learning schemes, with a similar total number of turns as a pure stereo vision run. The intuition behind this method being best is that it allows the drone to best learn from samples when the drone is beyond the normal stereo vision turning threshold. The original DAgger scheme has a larger probability to turn earlier, exploring these samples to a lesser extent. Double-sided statistical bootstrap tests \cite{cohen1996empirical} indicate that all differences between the learning methods are significant with $p < 0.05$. 

\begin{table}[]
    \centering
    \caption{Test results for the three learning schemes. The average and standard deviation are given for the number of overrides and turns during the testing period. A lower number of overrides is better. In the table, the best results are shown in bold.}\label{tab:results_simulation}
    \begin{tabular}{|l|c|c|}
    \hline
         Method & Overrides & Turns \\
         \hline
         Pure stereo & N/A & $45.6$ ($\sigma = 3.0$ )\\
         \hline
         1. Cold turkey & $25.1$ ($\sigma = 8.2 $ ) & $42.8$ ($\sigma = 3.7$ )\\
         \hline
         2. DAgger & $10.7$ ($\sigma = 5.3$ ) & $41.4$ ($\sigma = 3.2$ )\\
         \hline
         3. Training wheels & $\mathbf{4.3}$ ($\sigma = 2.6$ ) & $40.4$ ($\sigma = 2.6$ )\\
         \hline
    \end{tabular}
\end{table}

\begin{figure}
\centering
\includegraphics[width=8cm]{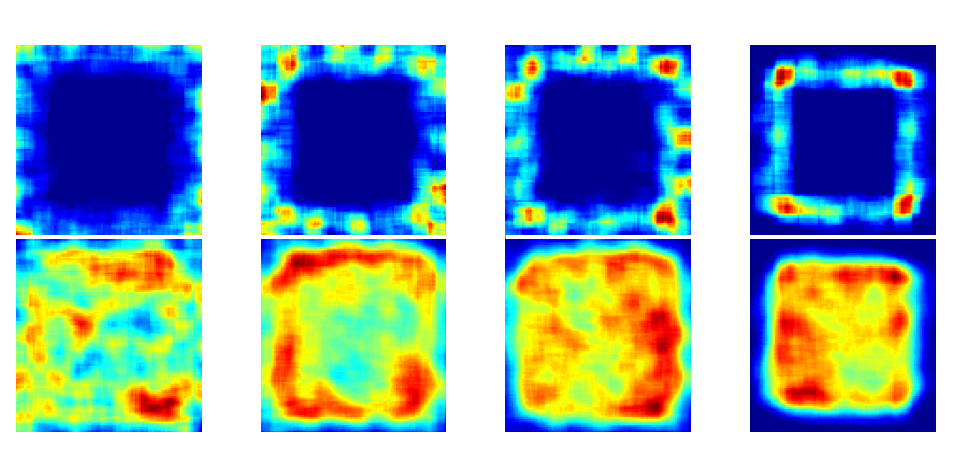}
\caption{Simulation heatmaps, from left to right: cold turkey, DAgger, training wheels, stereo only. Top images are turn locations, lower images are the approaches. }
\label{fig:simHeatmaps}
\end{figure}

The differences between the learning schemes are well illustrated by the positions the drone visits in the room during the test phase. Figure \ref{fig:simHeatmaps} contains `heat maps' that show the drone positions during turning (top row) and during straight flight (bottom row). The position distribution has been obtained by binning the positions during the test phase of all 30 runs. The results for each scheme are shown per column in Figure \ref{fig:simHeatmaps}. Right is the pure stereo vision scheme, which shows a clear border around the straight flight trajectories. It can be observed that this border is best approximated by the `training wheels' scheme (second from the right).   


\section{Robotic Experiments} \label{section:robot_experiments}
The simulation experiments showed that the `training wheels' setup resulted in the fewest stereo vision overrides when switching to monocular disparity estimation and control. In this section, we test this online learning setup with a flying robot.   

\begin{figure}
\centering
\includegraphics[width=4cm]{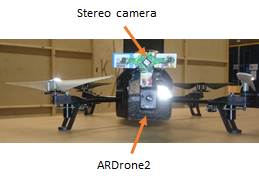}
\caption{The used multicopter.}
\label{fig:ardrone}
\end{figure}

The experiment is set up in the same manner as the simulation. The robot, a Parrot AR drone 2, first explores the room with the help of stereo vision. After 1 minute of learning, the drone switches to using the monocular disparity estimates with stereo vision running in the background for performing potential safety overrides. In this phase the drone still continues to learn. After learning 4 to 5 minutes, the drone stops learning and enters the test phase. Again, also for the real robot the main performance measure consists of the number of safety overrides performed by the stereo vision during the testing phase. 


The AR drone 2 is standard not equipped with a stereo vision system. Therefore, an in-house-developed 4 gram stereo vision system is used \cite{de2014autonomous}, which sends the raw images over USB to the ARDrone2. The grayscale stereo camera has a resolution of $128 \times 96$ px and is limited to $10$ fps. The ARDrone2 comes with a 1GHz ARM cortex A8 processor and 128MB RAM, and normally runs the Parrot firmware as an autopilot. For the experiments, we replace this firmware with the the open source Paparazzi autopilot software \cite{remes2013paparazzi,Brisset2014}. This allowed us to implement all vision and learning algorithms on board the drone. The length of each test is dependent on the battery, which due to wear has considerable variation, in the range of 8-15 minutes. 

\begin{figure}[htp]
\centering
\includegraphics[width=8.5cm]{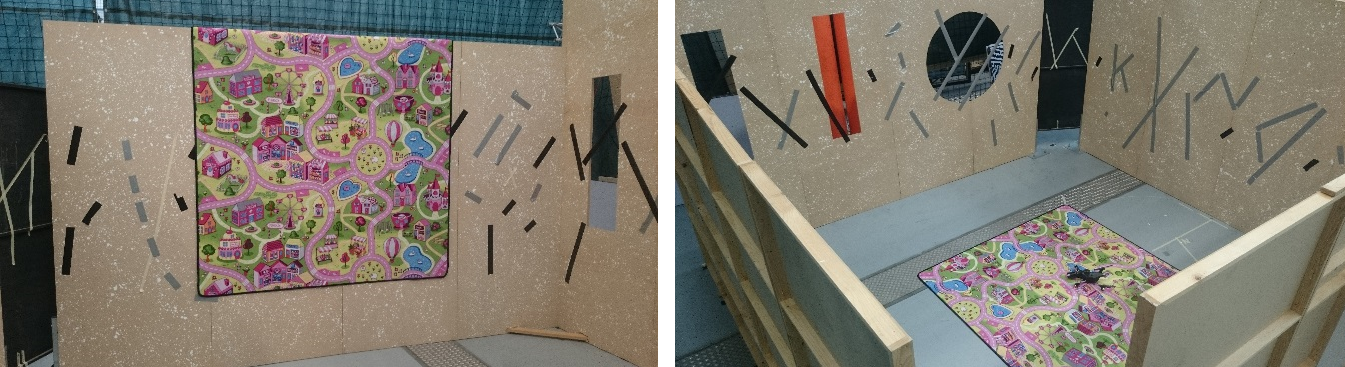}
\caption{Two test flight rooms.}
\label{fig:Rooms_overview}
\end{figure}

The tests are performed in an artificial room that has been constructed within a motion tracking arena. This allows us to track the trajectory of the drone and facilitates post-experiment analysis. The room is approximately $5 \times 5$ m, as delimited by plywood walls. In order to ensure that the stereo vision algorithm gave reliable results, we added texture in the form of duct-tape to the walls. In five tests, we had a textured carpet hanging over one of the walls (Figure \ref{fig:Rooms_overview} left, referred to as `room 1'), in the other five tests it was on the floor (Figure \ref{fig:Rooms_overview} right, referred to as `room 2'). 

\subsubsection{Results}
\begin{figure}[htp]
\centering
\includegraphics[trim={5.0cm 10.0cm 4.0cm 10.0cm},clip,width=8cm]{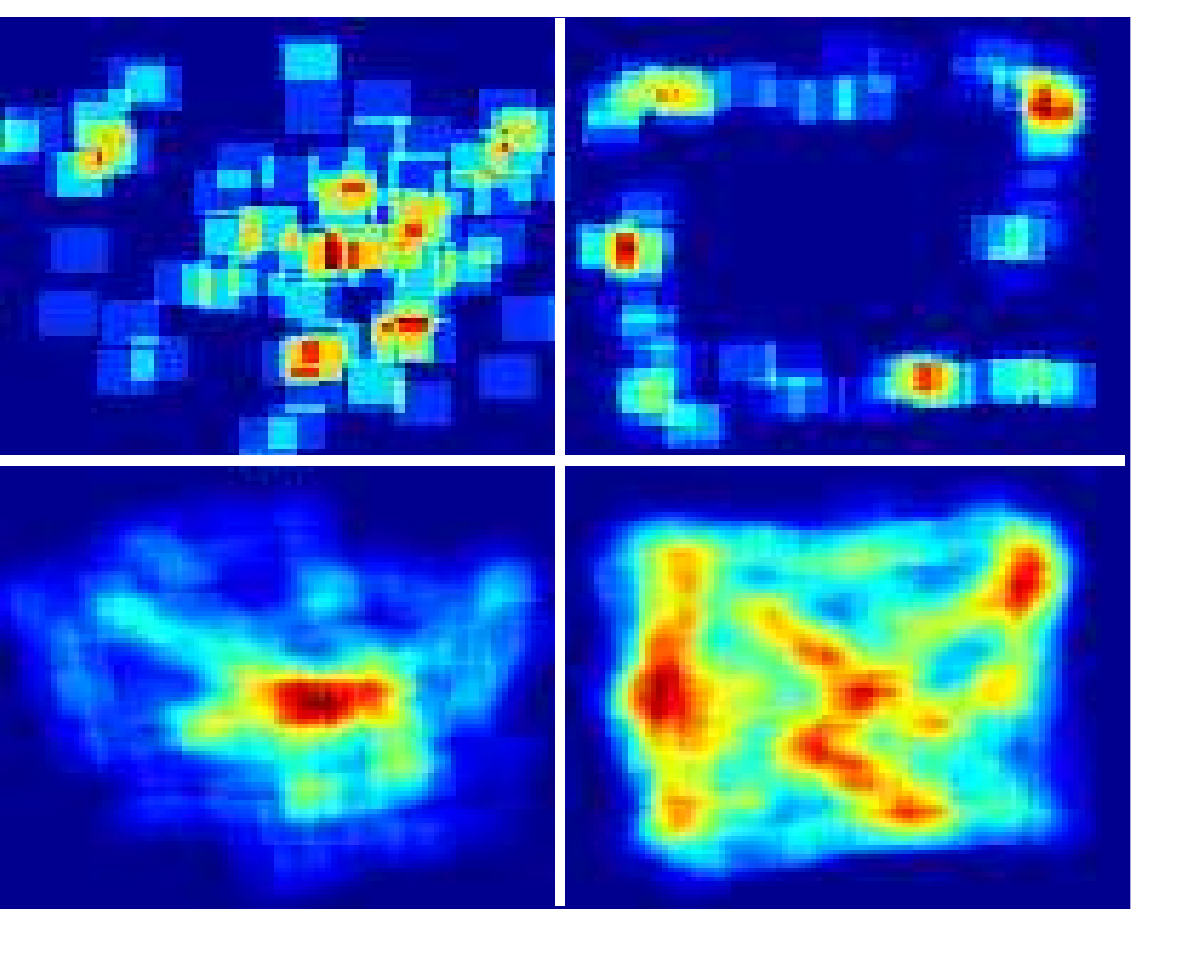}
\caption{Room 1 (plane texture) position heat map. Top row is the binned position during the avoidance turns, bottom row during the obstacle approaches, left column during stereo ground truth based operation, right column during learned monocular operation. }
\label{fig:heatmap_tapeRoom}
\end{figure}

Table \ref{table:2} shows the summarized results obtained from the monocular test flights. Two main observations can be made from this table. First, the average number of stereo overrides during the test phase is $3$, which is very close to the number of overrides in simulation. The monocular behavior also has a similar heat map to simulation. Figure \ref{fig:heatmap_tapeRoom} shows a heat map of the drone's position during the approaches and the avoidance maneuvers (the turns). Again, the stereo based flight performs better in the sense that the drone explores the room much more thoroughly and the turns happen consistently just before an obstacle is detected. On the other hand, especially in room 2, the monocular performance is quite good in the sense that the system is able to explore most of the room.

Second, the selected TPR and FPR are on average $0.47$ and $0.11$. The TPR is rather low compared to the offline tests. However, this number is heavily influenced by the monocular estimator based behavior. Due to the goal of the robot, avoiding obstacles slightly before the stereo ground truth recognizes them as positives, positives should hardly occur at al. Only in cases of FNs where the estimator is slower or wrong, positives will be registered by the ground truth. Similarly, the FPR is also lower in the context of the estimate based behavior. 

\begin{figure}[htp]
\centering
\includegraphics[trim={5.0cm 10.0cm 4.0cm 10.0cm},clip,width=8cm]{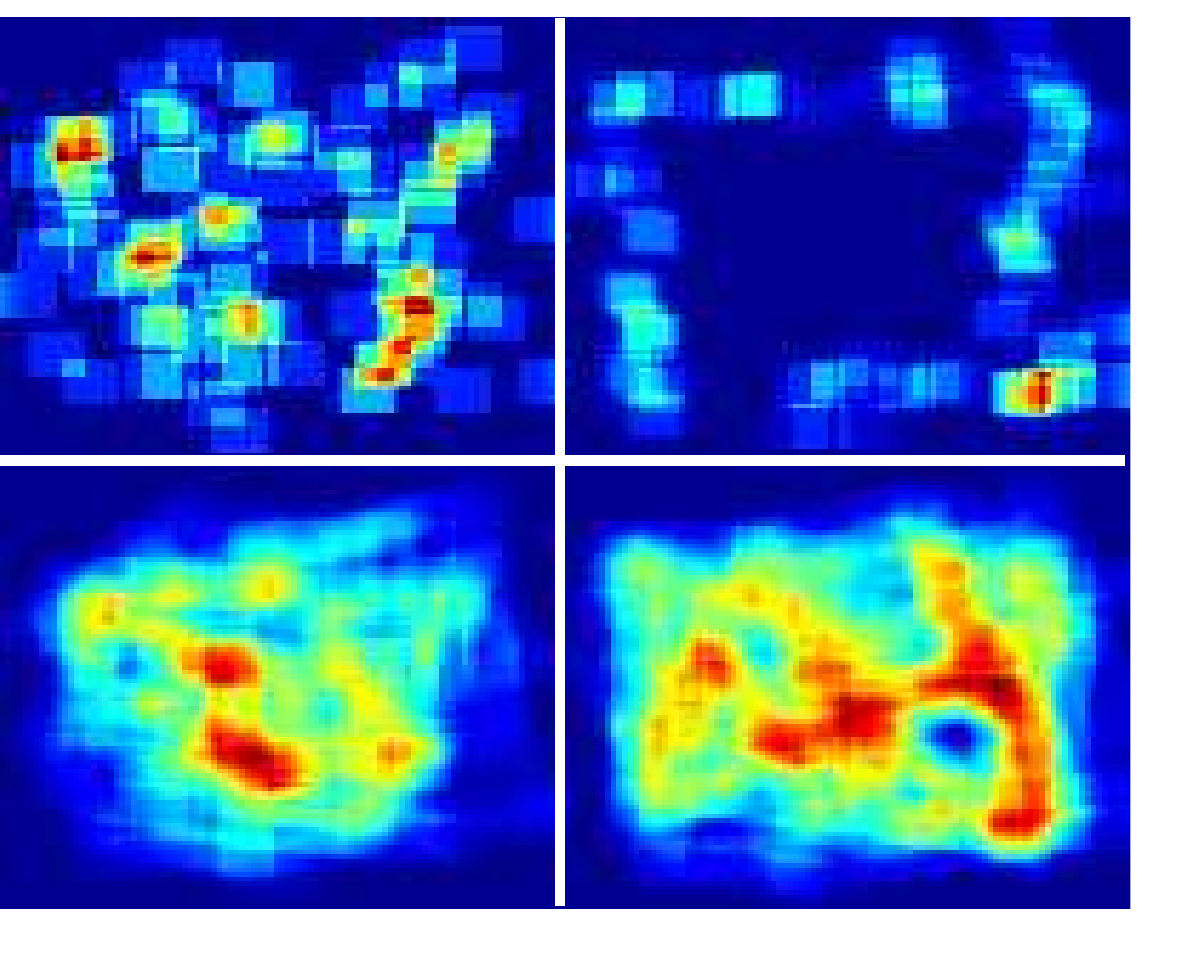}
\caption{Room 2 (carpet natural texture) position heat map.}
\label{fig:heatmap_carpetRoom}
\end{figure}

\begin{figure}[htp]
\centering
\includegraphics[trim={4.0cm 10.0cm 4.0cm 10.0cm},clip,width=8cm]{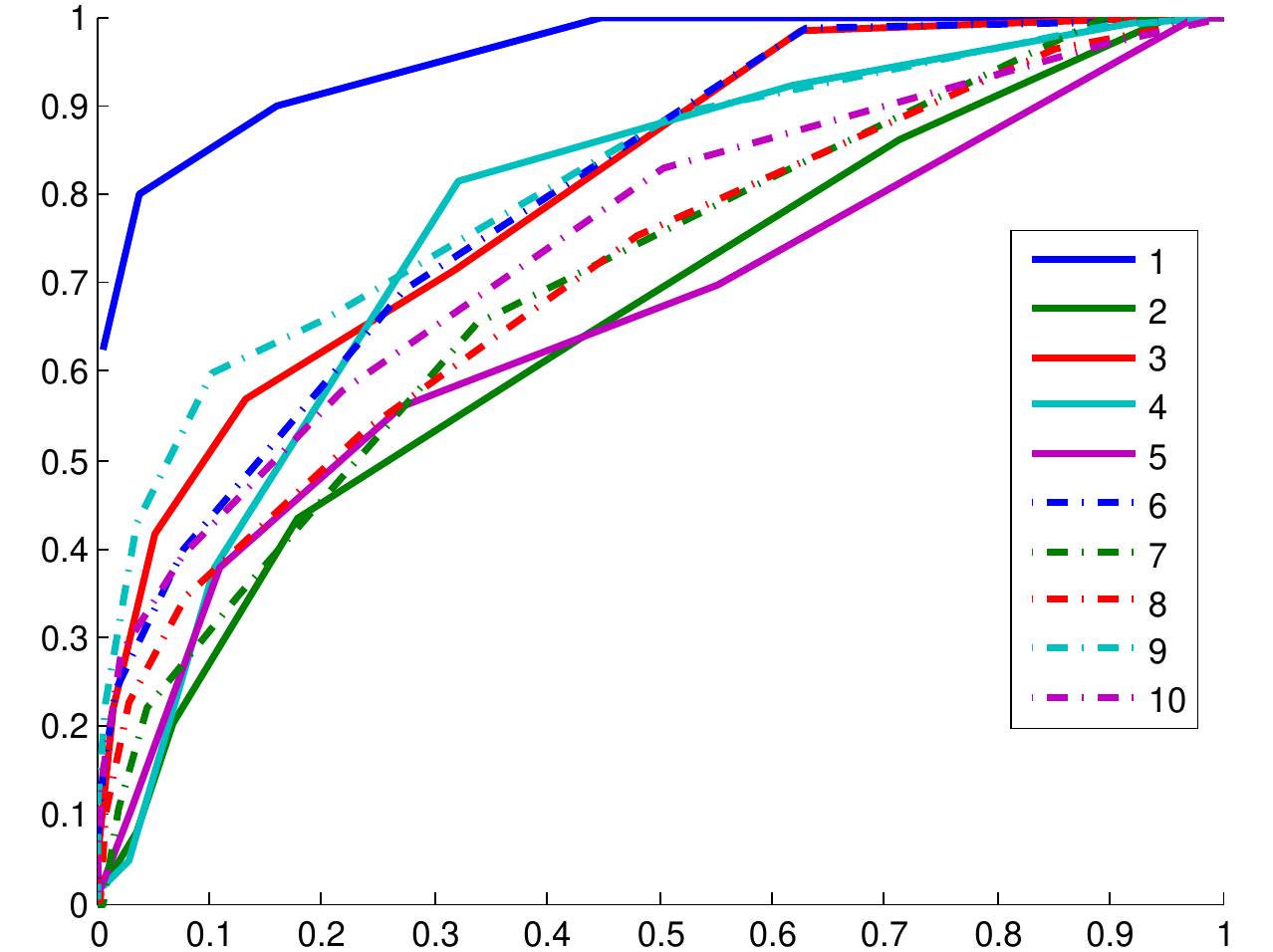}
\caption{ROC curves of the 10 test flights. Dashed/ solid lines refer to results on room \#1 / \#2.}
\label{fig:ROCCurves}
\end{figure}

ROC curves of the 10 flights are shown in Figure \ref{fig:ROCCurves}. A comparison based on the numbers between the first 5 flights (room 1) and the last 5 flights (room 2) does not show any significant differences, leading to the suggestion that the system is able to learn both rooms equally well. However, when comparing the heat maps of the two situations in the monocular system in Figure \ref{fig:heatmap_tapeRoom} and Figure \ref{fig:heatmap_carpetRoom}, it seems that the system shows slightly different behavior. The monocular system appears to explore room 2 better, getting closer to copying the behavior of the stereo based system. 


\begin{table*}[t]
\caption{Test flight summary}
\centering
\begin{tabular}{||c|c|c|c|c|c|c|c|c|c|c|c||} 
\hline
 & \multicolumn{5}{|c|}{Room 1} & \multicolumn{5}{|c|}{Room 2} & \\
 \hline
 Description & \#1 & \#2 & \#3 & \#4 & \#5 & \#6 & \#7 & \#8 & \#9 & \#10 & Avg.  \\ [0.5ex] 
 \hline
 Stereo flight time m:ss  & 6:48	 & 7:53	 & 2:13	 & 3:30	 & 4:45	 & 4:39	 & 4:56	 & 5:12	 & 4:58	 & 5:01	 & 4:59\\
 \hline
 Mono flight time m:ss & 3:44	 & 8:17	 & 6:45	 & 7:25	 & 4:54	 & 10:07	 & 4:46	 & 9:51	 & 5:23	 & 5:12	 & 6:39\\
 \hline
 Mean Square Error & 0,7	& 1,96	& 1,12	& 0,95	& 0,83	& 0,95	& 0,87	& 1,32	& 1,16	& 1,06	& 1,09\\
 \hline
 False Positive Rate & 0,16	& 0,18	& 0,13	& 0,11	& 0,11	& 0,08	& 0,13	& 0,08	& 0,1	& 0,08	& 0,11\\
 \hline
 True Positive Rate & 0,9	& 0,44	& 0,57	& 0,38	& 0,38	& 0,4	& 0,35	& 0,35	& 0,6	& 0,39	& 0,47\\
 \hline
Stereo approaches & 29	& 31	& 8	& 14	& 19	& 22	& 22	& 19	& 20	& 21	& 20,5 \\
\hline
Mono approaches & 10	& 21	& 20	& 25	& 14	& 33	& 15	& 28	& 18	& 15	& 19,9\\
\hline
Auto-overrides & 0	& 6	& 2	& 2	& 1	& 5	& 2	& 7	& 3	& 2	& 3\\
 \hline
Overrides ratio & 0 & 0.72 & 0.3 & 0.27 & 0.2 & 0.49 & 0.42 & 0.71 & 0.56 & 0.38 & 0.41 \\
 \hline
\end{tabular}
\label{table:2}
\end{table*}

\smallskip

\begin{figure*}
\centering
\includegraphics[trim={0cm 14cm 0cm 0cm},clip,width=18cm]{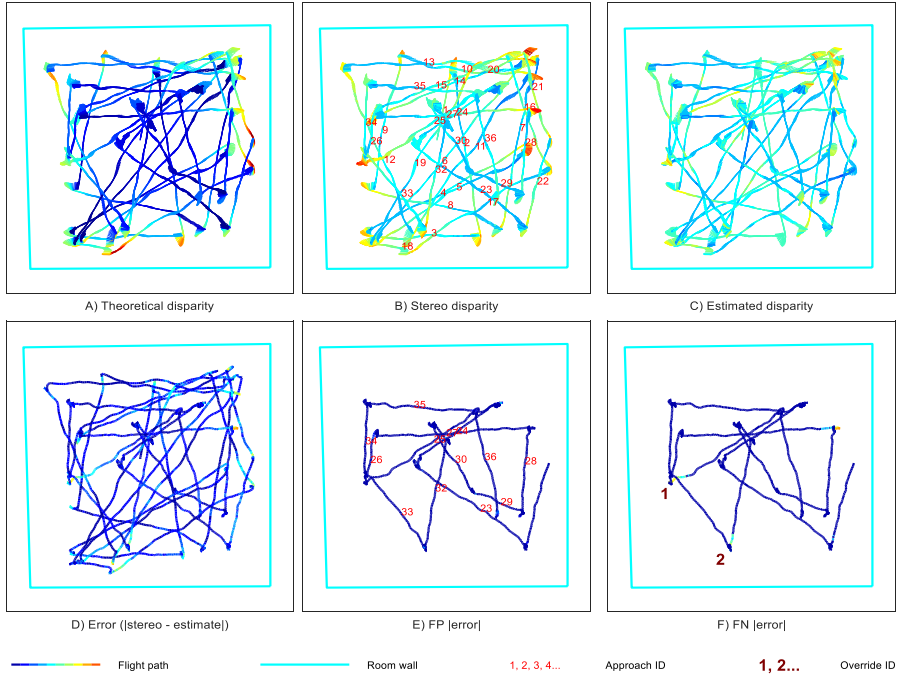}
\caption{Flight path of test 10 in room 2. Monocular flight starts form approach 23. The meaning of the color of the flightpath differs per image; A: the approximated disparity based on the external tracking system. B: the measured stereo average disparity, C: the monocular estimated disparity, D: the error between B\&C with dark blue meaning zero error, E: error during FP, F: error during FN. E and F only show the monocular part of the flight.}
\label{fig:flightpath10}
\end{figure*}

\smallskip
The experimental setup with the room in the motion tracking arena allows for a more in-depth analysis of the performance of both stereo and monocular vision. Figure \ref{fig:flightpath10} shows the spatial view of the flight trajectory of test \#10 \footnote{Onboard video data of the flight \#10 can be viewed at: \url{http://1drv.ms/1KC81PN}, an external video at: \url{https://youtu.be/lC_vj-QNy1I}, and a VBoW visualization video at: \url{http://1drv.ms/1fzvicH}}. The flight is segmented into approaches and turns which are numbered accordingly in these figures. The color scale in Figure \ref{fig:flightpath10}A is created by calculating the theoretically visible closest wall based on the tracking the system’s measured heading and position of the drone, the known position of the walls and the FOV angle of the camera. It is clearly visible that the stereo ground truth in Figure \ref{fig:flightpath10}B does not capture this theoretical disparity perfectly. Especially in the middle of the room the disparity remains high compared to the theoretical ground truth due to noise in the stereo disparity map. The results of the monocular estimator in Figure \ref{fig:flightpath10}C shows another decrease in quality compared to the stereo ground truth.  



\section{Discussion} \label{section:discussion}
We start the discussion with an interpretation of the results from the simulation and real-world experiments, after which we proceed by discussing persistent SSL in general and provide a comparison to other machine learning techniques.

\subsection{Interpretation of the results}
Using persistent SSL we were able to autonomously navigate our multicopter on the basis of a stereo vision camera, while training a monocular estimator on board and online. Although the monocular estimator allows the drone to continue flying and avoiding obstacles, the performance during the approximately ten-minute flights is not perfect. During monocular flight, a fairly limited amount of (autonomous) stereo overrides was needed while at the same time the robot was not fully exploring the room like when using stereo. 

Several improvements can be suggested. First, we can simply have the drone learn for a longer time, accumulating training data over multiple flights. In an extended offline test our VBoW method shows saturation at around 6000 samples. Using additional features and more advanced learning methods may result in improved performance if training set sizes increase.


During our tests in different environments, it proved unnecessary to tune the VBoW learning algorithm parameters to a new environment as similar performance was obtained. The learned results on the robot itself may or may not generalize to different environments; however, this is of less concern as the robot can detect a new environment and then decide to continue the learning process if the original cue is still available. In order to detect an inadequacy of the learned regression function, the robot can occasionally check the estimation error against the stereo ground truth. In fact our system already does so autonomously using its safety override. Methods on checking the performance without using the ground truth, e.g. by employing a learner that gives an estimate of uncertainty, are left for future work.

\subsection{Deep learning}
At the time of our robotic experiments, implementing state-of-the-art deep learning methods on-board a flying drone was deemed infeasible due to hardware restrictions. However, in Appendix \ref{section:deep_NN} we investigate using downsized networks in order to show that the persistent SSL method can work with deep learning as well. One of the major advantages of persistent SSL is the unprecedented amount of available training data. This amount of data will be more useful to more complex learning methods such as deep learning methods than to less complex, but computationally efficient methods such as the VBoW method used in our experiments. Today, with the availability of strongly improved hardware such as the NVidia Jetson TX1, close-to state-of-the-art models can be trained and run on-board a drone, which may significantly improve the learning results.

\subsection{Persistent SSL in relation to other machine learning techniques}
\begin{figure}[htp]
\centering
\includegraphics[trim={0cm 9cm 18cm 0cm},clip,width=8.5cm]{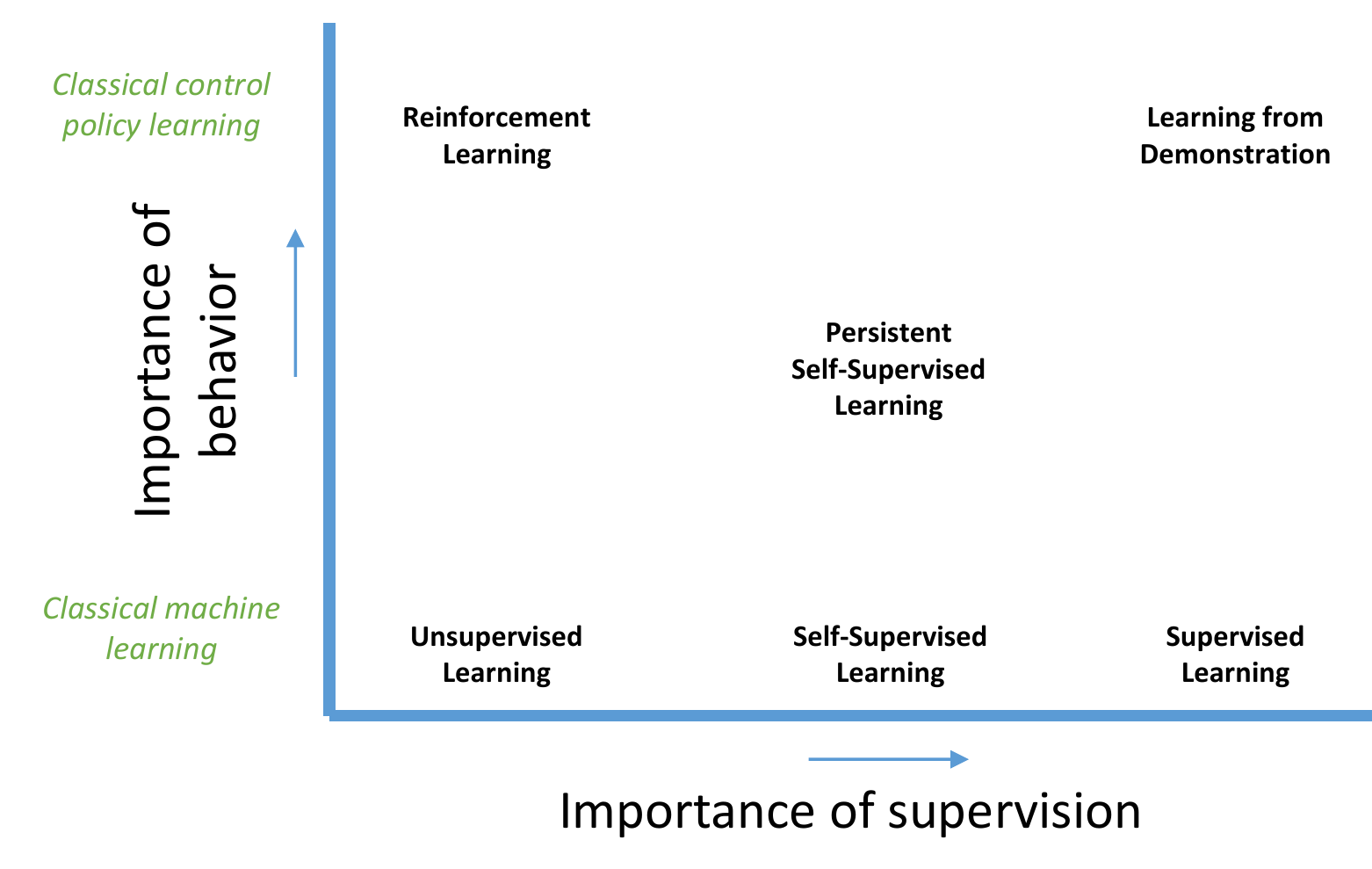}
\caption{Lay of the machine learning land}
\label{fig:LayOfMLLand}
\end{figure}

In order to place persistent SSL in the general framework of machine learning, we compare it with several techniques. An overview of this comparison is presented in figure \ref{fig:LayOfMLLand}.
\smallskip

\paragraph*{Un-/semi-/super-vised learning}
Unsupervised learning does not require labeled data, semi-supervised learning requires only an initial set of labeled data \cite{Zhu2007}, and supervised learning requires all the data to be labeled. Internally, persistent SSL uses a standard supervised learning scheme, which greatly facilitates and speeds up learning. The typical downside of supervised learning - acquiring the labels - does not apply to SSL, since the robot continuously provides these labels itself. 

A major difference between the typical use of supervised learning and its use in persistent SSL, is that the data for learning is generally assumed to be i.i.d. However, persistent SSL controls a behavioral component which, in turn, affects both the dataset obtained during training as well as during testing time. Operation based on ground truth induces a certain behavior that differs significantly from behavior induced from a trained estimator, even more so for an undertrained estimator.


\paragraph*{SSL}
The persistent form of SSL is set apart in the figure from `normal' SSL, because the persistence property introduces a much more significant behavioral component to the learning. While normal SSL expects the trusted cue to remain available, persistent SSL assumes that the robot may sometimes act in the absence of the trusted cue. This introduces the feedback-induced data bias problem, which, as we have seen, requires specific behavior strategies for best learning the robot's task.


\paragraph*{Learning from demonstration}
Learning from Demonstration (LfD), or Learning from Demonstration (LfD), is a close relative to persistent SSL. Consider for instance teleoperation, an LfD scheme in which a (human or robot) teacher remotely operates a robot in order for it to learn demonstrated actions in its environment \cite{Argall2009}. This can be compared to persistent SSL if we consider the teacher to be the ground truth function $g(x_g)$ in the persistent SSL scheme. In most cases described in literature, the teacher shows actions from a control policy taken on the basis of a state instead of just the results from a sensory cue (i.e., the state). However, LfD does contain exceptions in which the learner only records the states during demonstration, e.g. when drawing a map through a 2D representation of the world in case of a path planning mission in an outdoor robot \cite{Ratliff2006}. Like persistent SSL, test time decisions taken in LfD schemes influence future observations which may or may not be contained in known demonstrated territory. However, one key difference between LfD and persistent SSL arguably sets them apart. All LfD theory known to the authors implicitly assumes the teacher is never the same entity as the learner. It may be that all relevant sensors are on the learner, and even that the learner’s body is used to execute teacher commands (like in teleoperation), but the teacher’s intelligence is always an external entity.

\paragraph*{Reinforcement learning}
Lastly we compare persistent SSL with Reinforcement Learning (RL), which is a distinctively different technique \cite{Kober2013}. In RL a policy is learned using a reward function. Due to the evaluative feedback provided in RL, defining a good reward function is one of fundamental difficulties of RL known as reward shaping \cite{Littman2015}, \cite{Kober2013}. Since persistent SSL uses supervised feedback, reward shaping is less of an issue in persistent SSL, only requiring a choice of a loss function between $g(x_g)$ and $f(x_f)$. Secondly, the initial exploration phase of RL often infers a lot of trial-and-error, making it a dangerous time in which a physical system may crash and be damaged. Although this particular problem is often solved by better initialization, e.g. by using for instance LfD or using policy search instead of value function based approaches, persistent SSL does not require an untrained initialization phase at all as a reliable ground truth function guarantees a certain minimal correct behavior.

Persistent SSL differs from other learning techniques in the sense that no complete training data set is needed to train the algorithm beforehand. Instead it requires a ground truth $g(x_g)$, which must be available online in real-time while training $\hat{f}(x_f)$, but can be switched off when $\hat{f}(x_f)$ is learned to satisfaction. This implies that learning needs to be persistent and that the switch $\theta$ must be included in the model. Note that in cases where the environment of the robot may change, measures can be put in place to detect the output uncertainty of $\hat{f}(x_f)$ . If the uncertainty goes up, the robot can switch back to using the ground truth function and learning can then be activated again. Developing such measures is, however, left for future work.


\subsection{Feedback-induced data bias} 
The robot induces how its environment is perceived, meaning it influences the acquired training samples based on its behavior. The problems arising from this feedback-induced data bias are known from other machine learning disciplines, such as RL and LfD \cite{Kober2013}. In particular, Ross et al have proposed DAgger\cite{Ross2011} to solve a similar problem in the LfD domain, which iteratively aggregates the dataset with induced training samples and the expert’s reaction to it. However, in the case of LfD, obtaining the induced training samples requires a careful and often additional setup, while in persistent SSL this functionality is inherently available. Secondly the performance of the LfD expert (i.e. in many cases a human) is not easy to control, often reacting too late or too early. The control policy of the persistent SSL ground truth override system can, on the other hand, be very deterministic. In the case of a DAgger application with drones flying through a forest \cite{Ross2013a}, it proved infeasible to reliably sample the expert in an online fashion. Acquired videos had to be processed offline by the expert, hence the need for (offline $\leftrightarrow$ online) iterations. Moreover an additional online human safety override interface was still necessary to prevent damage to the drone while learning. Thirdly, due to the cost of (and need for) iterative demonstration sessions, the emphasis of DAgger is on converging fast with needing as little expert sessions as possible. In persistent SSL there are no costs for using the teacher signals coming from the original sensor cue. With persistent SSL we can directly focus on effectively using the available amount of training samples instead of minimizing the number of iterations like in DAgger.

Another reason why persistent SSL handles the induced training sample issue better than other state of the art robot learning methods is that in persistent SSL part of the learning problem itself can be easily separated and tested from the behavior; i.e. in a traditional supervised learning setting. In our proof of concept this has allowed us to test the learning algorithms and thoroughly investigate its limits before deployment, which helped us to identify when the behavioral influence was to blame for bad results. 

\section{Conclusion} \label{section:conclusion}

We have investigated a novel Self-Supervised Learning scheme, in which the supervisory signal is switched off after an initial learning period. In particular we have studied an instance of such persistent SSL for the task of obstacle avoidance, in which the robot uses trusted stereo vision distance estimates in order to learn appearance-based monocular distance estimation. We have shown that this particular setup is very similar to learning from demonstration. This similarity has been corroborated by experiments in simulation, which showed that the worst learning strategy is to make a hard switch from stereo vision flight to mono vision flight. It is best to have the robot fly based on mono vision and using stereo vision only as `training wheels', to take over when the robot would otherwise collide with an obstacle. The real-world robot experiments show the feasibility of the approach, giving acceptable results already with just 4-5 minutes of learning. 

The findings also indicate interesting future venues of investigation. First, and perhaps most importantly, in the 4-5 minutes of the real-world experiments the robot already experiences roughly 7000 - 9000 supervised learning samples. It is clear that longer learning times can lead to very large supervised data sets, which are suitable for deep learning approaches. Such approaches likely allow the learning to extend to much larger, more varied environments. In addition, they could allow the learning to improve the resolution of disparity estimates from a single value to a full image size disparity map. Second, in the current experiments the robot stayed in a single environment. We mentioned that a different environment can make the learned mapping invalid, and that this can be detected by means of the ground truth. Another venue, as studied in \cite{ho2015optical}, is to use a machine learning method with an associated uncertainty value. For instance, one could use a learning method such as a Gaussian Process. This can help with a further integration of the behavior with learning, for instance by tuning the forward velocity based on the certainty. These venues together could allow for persistent SSL to reach its full potential, significantly enhancing the robustness of robots operating in real-world environments.

\appendices
\section{Deep neural network results} \label{section:deep_NN}

\begin{figure*}
\centering
\includegraphics[trim={2cm 13cm 3cm 3cm},clip,width=18cm]{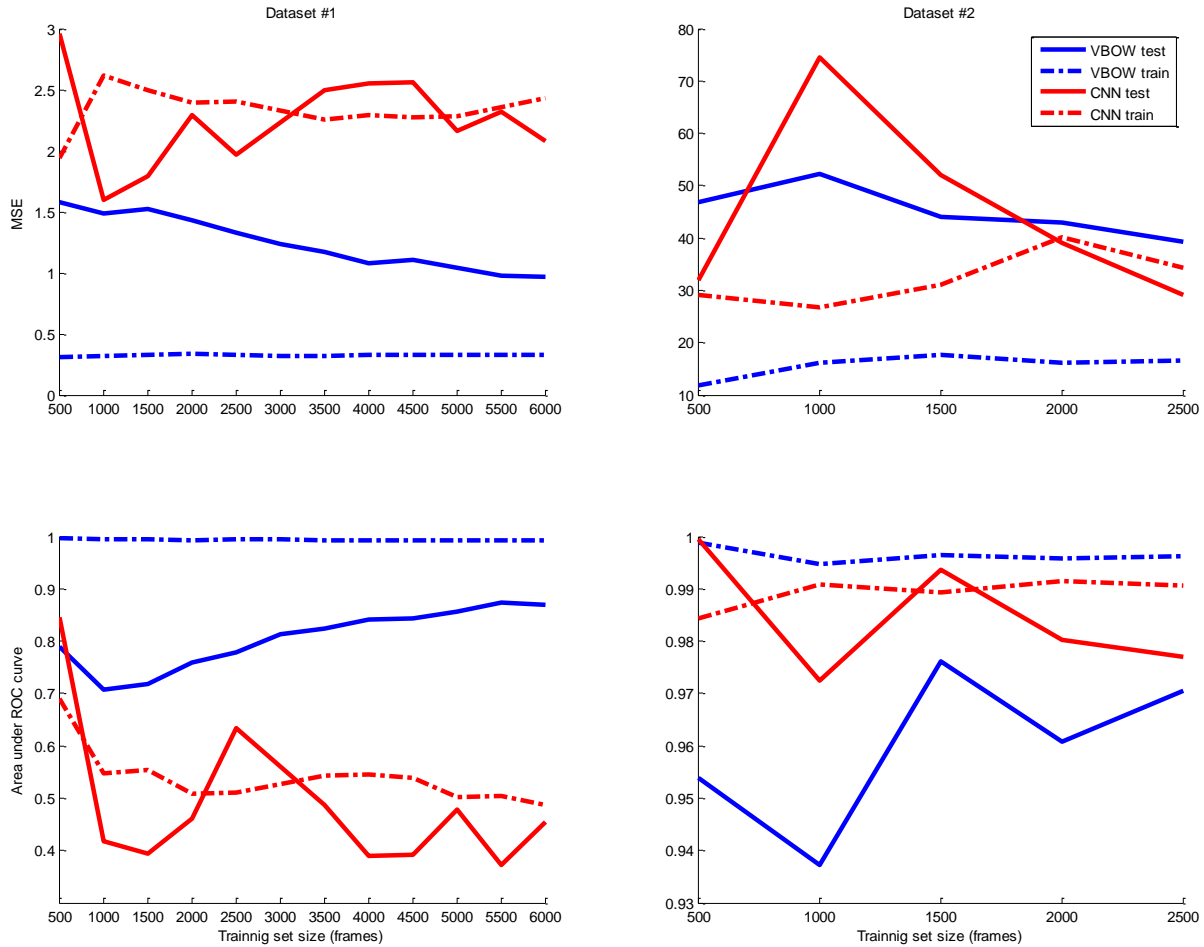}
\caption{Learning curves on two datasets for CNNs versus VBoW. Dashed/ solid lines refer to results on train/test set.}
\label{fig:CNNvsVBoW}
\end{figure*}

We investigated the possibility of implementing a deep convolutional neural network (CNN) on-board a drone to test our proposed learning scheme. We scaled the CNN architecture so that implementing this network on the ARDrone 2 hardware remained feasible. 
Due to CPU restrictions of our target system, we choose to use a relatively small CNN inspired by the Cifar10 CNN example used in Caffe. Our layers are as follows. Input: 128x128 pixels image (input images are scaled). First hidden layer: 5x5 convolution, max pooling to 3x3, 32 kernels, contrast normalization. Second hidden layer: 5x5 convolution, average pooling to 3x3, 32 kernels, contrast normalization. Third hidden layer: 5x5 convolution, average pooling to 3x3, 128 kernels. Fourth and fifth layer: fully connected. Lastly, an euclidean loss output layer to one single output. Furthermore, we used a learning rate of 1e-6, 0.9 momentum and used 10.000 iterations. The networks are trained using the open source Caffe framework on a dedicated compute server with an NVidia GTX660. 
In Figure  \ref{fig:CNNvsVBoW}, the learning curves of our best CNN under these constraints versus the best of our VBoW trained algorithms are shown. 
The figure shows that the CNN is able to learn the problem of monocular depth estimation to a comparable fasion as our VBoW method. For the expected amount of training data during a single flight, the VBoW method delivers better performance for dataset \#1 and slightly worse for dataset \#2.

\ifCLASSOPTIONcaptionsoff
  \newpage
\fi



\bibliographystyle{IEEEtran}
\bibliography{IEEEabrv,PSSL}

\begin{thebibliography}{10}
\providecommand{\url}[1]{#1}
\csname url@samestyle\endcsname
\providecommand{\newblock}{\relax}
\providecommand{\bibinfo}[2]{#2}
\providecommand{\BIBentrySTDinterwordspacing}{\spaceskip=0pt\relax}
\providecommand{\BIBentryALTinterwordstretchfactor}{4}
\providecommand{\BIBentryALTinterwordspacing}{\spaceskip=\fontdimen2\font plus
\BIBentryALTinterwordstretchfactor\fontdimen3\font minus
  \fontdimen4\font\relax}
\providecommand{\BIBforeignlanguage}[2]{{%
\expandafter\ifx\csname l@#1\endcsname\relax
\typeout{** WARNING: IEEEtran.bst: No hyphenation pattern has been}%
\typeout{** loaded for the language `#1'. Using the pattern for}%
\typeout{** the default language instead.}%
\else
\language=\csname l@#1\endcsname
\fi
#2}}
\providecommand{\BIBdecl}{\relax}
\BIBdecl

\bibitem{garcia2015comprehensive}
J.~Garc{\'\i}a and F.~Fern{\'a}ndez, ``A comprehensive survey on safe
  reinforcement learning,'' \emph{Journal of Machine Learning Research},
  vol.~16, pp. 1437--1480, 2015.

\bibitem{Ross2011}
S.~Ross, G.~J. Gordon, and J.~A. Bagnell, ``{A Reduction of Imitation Learning
  and Structured Prediction},'' \emph{14th International Conference on
  Artificial Intelligence and Statistics}, vol.~15, 2011.

\bibitem{Ross2013a}
\BIBentryALTinterwordspacing
S.~Ross, N.~Melik-Barkhudarov, K.~S. Shankar, A.~Wendel, D.~Dey, J.~A. Bagnell,
  and M.~Hebert, ``{Learning monocular reactive UAV control in cluttered
  natural environments},'' \emph{2013 IEEE International Conference on Robotics
  and Automation}, pp. 1765--1772, May 2013. [Online]. Available:
  \url{http://ieeexplore.ieee.org/lpdocs/epic03/wrapper.htm?arnumber=6630809}
\BIBentrySTDinterwordspacing

\bibitem{thrun2007stanley}
S.~Thrun, M.~Montemerlo, H.~Dahlkamp, D.~Stavens, A.~Aron, J.~Diebel, P.~Fong,
  J.~Gale, M.~Halpenny, G.~Hoffmann \emph{et~al.}, ``Stanley: The robot that
  won the darpa grand challenge,'' in \emph{The 2005 DARPA Grand
  Challenge}.\hskip 1em plus 0.5em minus 0.4em\relax Springer, 2007, pp. 1--43.

\bibitem{lieb2005adaptive}
D.~Lieb, A.~Lookingbill, and S.~Thrun, ``Adaptive road following using
  self-supervised learning and reverse optical flow.'' in \emph{Robotics:
  Science and Systems}, 2005, pp. 273--280.

\bibitem{lookingbill2007reverse}
A.~Lookingbill, J.~Rogers, D.~Lieb, J.~Curry, and S.~Thrun, ``Reverse optical
  flow for self-supervised adaptive autonomous robot navigation,''
  \emph{International Journal of Computer Vision}, vol.~74, no.~3, pp.
  287--302, 2007.

\bibitem{Hadsell2009b}
\BIBentryALTinterwordspacing
R.~Hadsell, P.~Sermanet, J.~Ben, A.~Erkan, M.~Scoffier, K.~Kavukcuoglu,
  U.~Muller, and Y.~LeCun, ``{Learning long-range vision for autonomous
  off-road driving},'' \emph{Journal of Field Robotics}, vol.~26, no.~2, pp.
  120--144, 2009. [Online]. Available:
  \url{http://doi.wiley.com/10.1002/rob.20276
  http://onlinelibrary.wiley.com/doi/10.1002/rob.20276/abstract}
\BIBentrySTDinterwordspacing

\bibitem{muller2013real}
U.~A. Muller, L.~D. Jackel, Y.~LeCun, and B.~Flepp, ``Real-time adaptive
  off-road vehicle navigation and terrain classification,'' in \emph{SPIE
  Defense, Security, and Sensing}.\hskip 1em plus 0.5em minus 0.4em\relax
  International Society for Optics and Photonics, 2013, pp. 87\,410A--87\,410A.

\bibitem{baleia2015exploiting}
J.~Baleia, P.~Santana, and J.~Barata, ``On exploiting haptic cues for
  self-supervised learning of depth-based robot navigation affordances,''
  \emph{Journal of Intelligent \& Robotic Systems}, pp. 1--20, 2015.

\bibitem{ho2015optical}
H.~Ho, C.~De~Wagter, B.~Remes, and G.~de~Croon, ``Optical flow for
  self-supervised learning of obstacle appearance,'' in \emph{Intelligent
  Robots and Systems (IROS), 2015 IEEE/RSJ International Conference on}.\hskip
  1em plus 0.5em minus 0.4em\relax IEEE, 2015, pp. 3098--3104.

\bibitem{Mori2013}
T.~Mori and S.~Scherer, ``{First results in detecting and avoiding frontal
  obstacles from a monocular camera for micro unmanned aerial vehicles},''
  \emph{Proceedings - IEEE International Conference on Robotics and
  Automation}, pp. 1750--1757, 2013.

\bibitem{Engel2014}
\BIBentryALTinterwordspacing
J.~Engel, J.~Sturm, and D.~Cremers, ``{Scale-aware navigation of a low-cost
  quadrocopter with a monocular camera},'' \emph{Robotics and Autonomous
  Systems}, vol.~62, no.~11, pp. 1646--1656, 2014. [Online]. Available:
  \url{http://www.sciencedirect.com/science/article/pii/S0921889014000566}
\BIBentrySTDinterwordspacing

\bibitem{engel2014scale}
------, ``Scale-aware navigation of a low-cost quadrocopter with a monocular
  camera,'' \emph{Robotics and Autonomous Systems}, vol.~62, no.~11, pp.
  1646--1656, 2014.

\bibitem{van2014monocular}
F.~van Breugel, K.~Morgansen, and M.~H. Dickinson, ``Monocular distance
  estimation from optic flow during active landing maneuvers,''
  \emph{Bioinspiration \& biomimetics}, vol.~9, no.~2, p. 025002, 2014.

\bibitem{de2016monocular}
G.~C. de~Croon, ``Monocular distance estimation with optical flow maneuvers and
  efference copies: a stability-based strategy,'' \emph{Bioinspiration \&
  biomimetics}, vol.~11, no.~1, p. 016004, 2016.

\bibitem{DeCroon2012b}
G.~de~Croon, M.~Groen, C.~{De Wagter}, B.~Remes, R.~Ruijsink, and B.~van
  Oudheusden, ``{Design, aerodynamics and autonomy of the DelFly},''
  \emph{Bioinspiration \& Biomimetics}, vol.~7, no.~2, p. 025003, 2012.

\bibitem{Argall2009}
\BIBentryALTinterwordspacing
B.~D. Argall, S.~Chernova, M.~Veloso, and B.~Browning, ``{A survey of robot
  learning from demonstration},'' \emph{Robotics and Autonomous Systems},
  vol.~57, no.~5, pp. 469--483, 2009. [Online]. Available:
  \url{http://dx.doi.org/10.1016/j.robot.2008.10.024}
\BIBentrySTDinterwordspacing

\bibitem{Hoiem2005}
D.~Hoiem, A.~a. Efros, and M.~Hebert, ``{Automatic photo pop-up},'' \emph{ACM
  Transactions on Graphics}, vol.~24, no.~3, p. 577, 2005.

\bibitem{Saxena2007a}
\BIBentryALTinterwordspacing
A.~Saxena, S.~H. Chung, and A.~Y. Ng, ``{3-D Depth Reconstruction from a Single
  Still Image},'' \emph{International Journal of Computer Vision}, vol.~76,
  no.~1, pp. 53--69, Aug. 2007. [Online]. Available:
  \url{http://link.springer.com/10.1007/s11263-007-0071-y}
\BIBentrySTDinterwordspacing

\bibitem{Saxena2009a}
\BIBentryALTinterwordspacing
A.~Saxena, M.~Sun, and A.~Y. Ng, ``{Make3D: learning 3D scene structure from a
  single still image.}'' \emph{IEEE transactions on pattern analysis and
  machine intelligence}, vol.~31, no.~5, pp. 824--40, May 2009. [Online].
  Available: \url{http://www.ncbi.nlm.nih.gov/pubmed/19299858}
\BIBentrySTDinterwordspacing

\bibitem{Lenz2012}
\BIBentryALTinterwordspacing
I.~Lenz, M.~Gemici, and A.~Saxena, ``{Low-power parallel algorithms for single
  image based obstacle avoidance in aerial robots},'' \emph{IEEE International
  Conference on Intelligent Robots and Systems}, pp. 772--779, 2012. [Online].
  Available:
  \url{http://ieeexplore.ieee.org/lpdocs/epic03/wrapper.htm?arnumber=6386146}
\BIBentrySTDinterwordspacing

\bibitem{Eigen2014}
\BIBentryALTinterwordspacing
D.~Eigen, C.~Puhrsch, and R.~Fergus, ``{Depth map prediction from a single
  image using a multi-scale deep network},'' \emph{Advances in Neural
  Information \ldots}, pp. 1--9, 2014. [Online]. Available:
  \url{http://papers.nips.cc/paper/5539-tree-structured-gaussian-process-approximations}
\BIBentrySTDinterwordspacing

\bibitem{Michels2005}
\BIBentryALTinterwordspacing
J.~Michels, A.~Saxena, and A.~Y. Ng, ``{High speed obstacle avoidance using
  monocular vision and reinforcement learning},'' in \emph{Proceedings of the
  22nd International Conference on Machine Learning}, vol.~3, no. Suppl
  1.\hskip 1em plus 0.5em minus 0.4em\relax ACM Press, 2005, pp. 593--600.
  [Online]. Available:
  \url{http://portal.acm.org/citation.cfm?doid=1102351.1102426$\backslash$nhttp://dl.acm.org/citation.cfm?id=1102426
  http://portal.acm.org/citation.cfm?id=1102426}
\BIBentrySTDinterwordspacing

\bibitem{Dey}
D.~Dey, K.~S. Shankar, S.~Zeng, R.~Mehta, M.~T. Agcayazi, C.~Eriksen,
  S.~Daftry, M.~Hebert, and J.~A. Bagnell, ``{Vision and Learning for
  Deliberative Monocular Cluttered Flight}.''

\bibitem{Yamauchi1999}
K.~Yamauchi, M.~Oota, and N.~Ishii, ``{A self-supervised learning system for
  pattern recognition by sensory integration},'' \emph{Neural Networks},
  vol.~12, no.~10, pp. 1347--1358, 1999.

\bibitem{Thrun2007a}
S.~Thrun, M.~Montemerlo, H.~Dahlkamp, D.~Stavens, A.~Aron, J.~Diebel, P.~Fong,
  J.~Gale, M.~Halpenny, G.~Hoffmann, K.~Lau, C.~Oakley, M.~Palatucci, V.~Pratt,
  P.~Stang, S.~Strohband, C.~Dupont, L.~E. Jendrossek, C.~Koelen, C.~Markey,
  C.~Rummel, J.~van Niekerk, E.~Jensen, P.~Alessandrini, G.~Bradski, B.~Davies,
  S.~Ettinger, A.~Kaehler, A.~Nefian, and P.~Mahoney, ``{Stanley: The robot
  that won the DARPA Grand Challenge},'' \emph{Springer Tracts in Advanced
  Robotics}, vol.~36, pp. 1--43, 2007.

\bibitem{Wagter2014a}
C.~De~Wagter, S.~Tijmons, B.~Remes, and G.~de~Croon, ``{Autonomous Flight of a
  20-gram Flapping Wing MAV with a 4-gram Onboard Stereo Vision System},'' in
  \emph{IEEE International Conference on Robotics \& Automation (ICRA)}, no.
  Section IV, 2014, pp. 4982--4987.

\bibitem{DeCroon2012c}
G.~C. H.~E. {De Croon}, E.~{De Weerdt}, C.~{De Wagter}, B.~D.~W. Remes, and
  R.~Ruijsink, ``{The appearance variation cue for obstacle avoidance},''
  \emph{IEEE Transactions on Robotics}, vol.~28, no.~2, pp. 529--534, 2012.

\bibitem{Varma2003}
M.~Varma and A.~Zisserman, ``{Texture Classification : Are Filter Banks
  Necessary ? 1 Introduction 2 A review of the VZ classifier},'' \emph{Computer
  Vision and Pattern Recognition 2003 Proceedings 2003 IEEE Computer Society
  Conference on}, 2003.

\bibitem{Wu2004}
B.~Wu, T.~L. Ooi, and Z.~J. He, ``{Perceiving distance accurately by a
  directional process of integrating ground information.}'' \emph{Nature}, vol.
  428, no. 6978, pp. 73--77, 2004.

\bibitem{de2007holiday50av}
C.~De~Wagter and M.~Amelink, ``Holiday50av technical paper,'' 2007.

\bibitem{cohen1996empirical}
P.~R. Cohen, ``Empirical methods for artificial intelligence,'' \emph{IEEE
  Intelligent Systems}, no.~6, p.~88, 1996.

\bibitem{de2014autonomous}
C.~De~Wagter, S.~Tijmons, B.~D. Remes, and G.~C. de~Croon, ``Autonomous flight
  of a 20-gram flapping wing mav with a 4-gram onboard stereo vision system,''
  in \emph{Robotics and Automation (ICRA), 2014 IEEE International Conference
  on}.\hskip 1em plus 0.5em minus 0.4em\relax IEEE, 2014, pp. 4982--4987.

\bibitem{remes2013paparazzi}
B.~Remes, D.~Hensen, F.~Van~Tienen, C.~De~Wagter, E.~Van~der Horst, and
  G.~De~Croon, ``Paparazzi: how to make a swarm of parrot ar drones fly
  autonomously based on gps,'' in \emph{IMAV 2013: Proceedings of the
  International Micro Air Vehicle Conference and Flight Competition, Toulouse,
  France, 17-20 September 2013}, 2013.

\bibitem{Brisset2014}
P.~Brisset, A.~Drouin, M.~Gorraz, P.-s. Huard, P.~Brisset, A.~Drouin,
  M.~Gorraz, P.-s. Huard, J.~T.~T. Pa, P.~Brisset, A.~Drouin, M.~Gorraz, P.-s.
  Huard, and J.~Tyler, ``{The Paparazzi Solution},'' 2014.

\bibitem{Zhu2007}
\BIBentryALTinterwordspacing
X.~Zhu, ``{Semi-Supervised Learning Literature Survey},'' \emph{Sciences-New
  York}, pp. 1--59, 2007. [Online]. Available:
  \url{http://citeseerx.ist.psu.edu/viewdoc/download?doi=10.1.1.146.2352\&rep=rep1\&type=pdf}
\BIBentrySTDinterwordspacing

\bibitem{Ratliff2006}
N.~Ratliff, D.~Bradley, J.~A. Bagnell, and J.~Chestnutt, ``{Boosting Structured
  Prediction for Imitation Learning for Imitation Learning},'' \emph{Advances
  in Neural Information Processing Systems (NIPS),}, p.~54, 2006.

\bibitem{Kober2013}
\BIBentryALTinterwordspacing
J.~Kober, J.~a. Bagnell, and J.~Peters, ``{Reinforcement learning in robotics:
  A survey},'' \emph{The International Journal of Robotics Research}, vol.~32,
  pp. 1238--1274, 2013. [Online]. Available:
  \url{http://ijr.sagepub.com/cgi/doi/10.1177/0278364913495721}
\BIBentrySTDinterwordspacing

\bibitem{Littman2015}
\BIBentryALTinterwordspacing
M.~L. Littman, ``{Reinforcement learning improves behaviour from evaluative
  feedback},'' \emph{Nature}, vol. 521, no. 7553, pp. 445--451, 2015. [Online].
  Available: \url{http://www.nature.com/doifinder/10.1038/nature14540}
\BIBentrySTDinterwordspacing

\end{thebibliography}
%
%
%

%





\end{document}